%% file: paper.tex
\documentclass[]{bytedance_seed}

\input{macros}

\title{\name{}: A Unified Model for Robot Interaction, Reasoning and Planning}

\author[*]{Huang Fang}
\author[*]{Mengxi Zhang}
\author[*]{Heng Dong}
\author[*\dagger]{Wei Li}
\author{\\Zixuan Wang}
\author{Qifeng Zhang}
\author{Xueyun Tian}
\author{Yucheng Hu}
\author[\dagger]{Hang Li}

\affiliation[]{ByteDance Seed}

\contribution[*]{Equal Contribution}
\contribution[\dagger]{Project Lead}

\abstract{

We introduce \textbf{\name{}}, a unified model that integrates robot reasoning, task planning, and natural language interaction within a single vision-language architecture.  
Acting as the high-level cognitive layer in a hierarchical robot system, \name{} dynamically generates atomic commands for the low-level controller and verbal responses for human interaction, enabling robots to follow complex instructions, plan long-horizon tasks, and interact naturally with human within an end-to-end framework. 
\name{} further introduces novel capabilities such as proactive dialogue, real-time interruption handling, and context-aware commonsense reasoning during task execution.
At its core, \name{} leverages chain-of-thought reasoning and adopts a three-stage training strategy: (1) continued pretraining to enhance foundational embodied reasoning abilities including 3D spatial understanding, visual grounding, and task-centric reasoning; (2) supervised finetuning to model human-robot interaction and task planning as a unified reasoning-action sequence; and (3) reinforcement learning to improve reasoning-action consistency and long-horizon task coherence.
Extensive experiments demonstrate that \name{} outperforms both open-source and commercial baselines (\eg{}, GPT-4o and Gemini 2.5 Pro) in interactive task execution, demonstrating strong generalization across diverse instruction types (\eg{}, open-ended, multi-stage, constrained, invalid, and interrupted) and various user-involved tasks such as table bussing, grocery shopping, and dietary filtering.
}

\date{\today}
\correspondence{\email{liwei.85@bytedance.com}, \email{lihang.lh@bytedance.com}}
\checkdata[Project Page]{\url{https://robix-seed.github.io/robix/}}

\begin{document}
\maketitle


\input{sections/1-introduction}
\input{sections/2-model}
\input{sections/3-data_training}
\input{sections/4-experiments}
\input{sections/5-relatedwork}
\input{sections/6-conclusion}
\input{sections/7-contribution}


\bibliographystyle{plainnat}
\bibliography{main}

\clearpage
\beginappendix
\input{sections/appendix}

\end{document}

%% file: macros.tex
\usepackage{minitoc}
\usepackage{url}
\usepackage{amssymb}
\usepackage[utf8]{inputenc}
\usepackage{microtype}
\usepackage{booktabs}
\usepackage{pifont} 
\usepackage{multirow}
\usepackage{makecell}
\usepackage{paralist}
\usepackage{xspace}
\usepackage{color}
\usepackage{xcolor}
\usepackage{colortbl}
\usepackage{adjustbox}
\usepackage{hyperref} 
\usepackage[edges]{forest}
\usepackage{tikz} 
\usepackage{caption}
\usepackage{amsfonts}
\usepackage[toc,page,header]{appendix}
\usepackage[utf8]{inputenc}
\usepackage{bbm}
\usepackage{enumitem}
\usepackage{booktabs}
\definecolor{seedc}{RGB}{7, 92, 173}
\usepackage{cleveref}
\usepackage{fancyvrb}
\usepackage{CJKutf8}
\usepackage{threeparttable}

\newcommand{\name}[1]{Robix}

\renewcommand{\paragraph}[1]{\vspace{0.1em}\noindent\textbf{#1}}

\newcommand{\ie}{\textit{i}.\textit{e}.}
\newcommand{\eg}{\textit{e}.\textit{g}.}

\usepackage[most]{tcolorbox} 
\usepackage{xcolor}          
\usepackage{lipsum}          
\definecolor{usercolor}{rgb}{0.88, 0.96, 1.0}  
\definecolor{robotcolor}{rgb}{0.95, 0.95, 0.95} 

\newlength{\dialogboxthickness}

%% file: sections/1-introduction.tex
\section{Introduction}
\label{sect:intro}

\begin{figure}[htbp]
    \centering
    \includegraphics[width=0.95\linewidth]{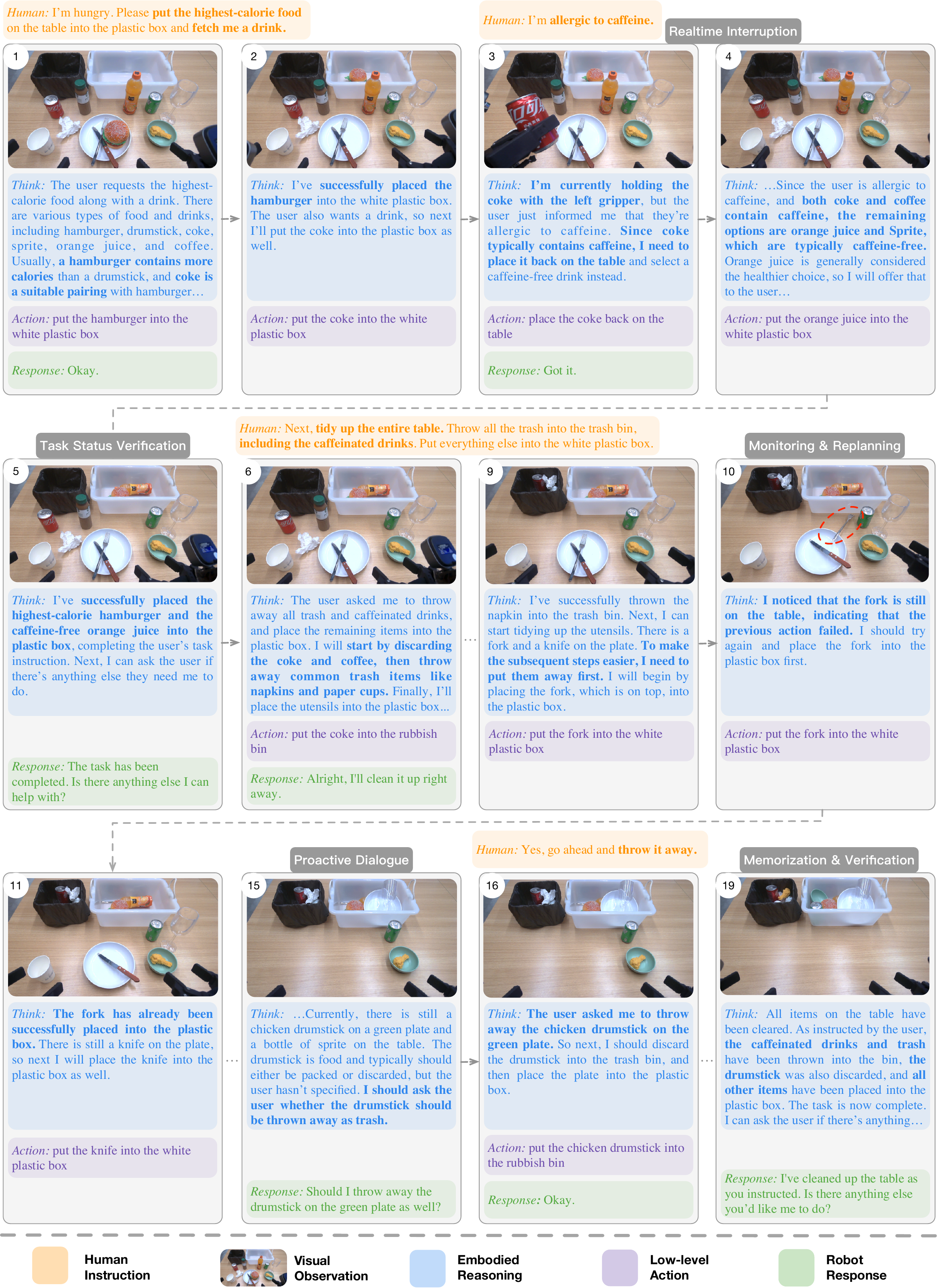}
    \caption{
        A demo of \name{}, showcasing (1) complex instruction understanding with commonsense reasoning; (2) real-time interruption handling; (3) task-status monitoring and dynamic replanning; and (4) proactive dialogue to clarify ambiguous instructions or infer user intent.
    }
    \label{fig:demo-cases}
\end{figure}

The goal of generalist robots is to assist humans in diverse daily tasks within open, dynamic environments. Realizing this vision requires more than executing isolated commands—it demands the ability to engage in natural human interaction and reason through complex, long-horizon tasks. For example, when cleaning a dining table, a robot must not only recognize dishes and utensils, but also interpret nuanced instructions such as ``\textit{only clear the plates if people are finished eating}'', respond to corrections like ``\textit{leave that glass}'', and adapt to novel scenarios such as properly sorting stacked tableware. 
To meet these requirements, a general-purpose robot system should adopt a hierarchical architecture in which a high-level cognitive layer handles complex multimodal reasoning, adaptive task planning, and natural human-robot interaction, while a low-level controller layer executes the atomic motor actions issued by the high-level layer. This division of responsibilities allows the robot to reason at a macro level while acting at a micro level, enabling human-like adaptability in real-world scenarios.

Existing hierarchical approaches typically employ large language models (LLMs) or vision language models (VLMs) as the high-level cognitive layer for task planning, which decompose long-horizon tasks into executable subtasks for the low-level controller~\cite{zhi2024closed,ahn2022can,huang2022inner,song2023llm,zhang2024hirt,cheng2024navila,duan2024manipulate,shah2024bumble}. However, these methods focus solely on task decomposition, overlooking human-robot interaction and embodied reasoning, which are essential for general-purpose robotic systems. 
Taking one step further, recent work \cite{colombani2024one} constructs modular pipelines that combine reasoning, planning, and interaction through hand-designed workflows. While workflow-based systems are easy to develop, their inflexibility and brittleness remain notable limitations—rooted primarily in rigid modularization and over-reliance on hand-engineered designs.
In this work, we introduce \textbf{\name{}}, a unified high-level cognitive layer that seamlessly integrates reasoning, task planning, and natural language interaction within a single model. Unlike modular frameworks, \name{} adopts an end-to-end vision–language architecture natively designed for interactive task execution.
At its core, \name{} leverages chain-of-thought reasoning and formulates interactive task execution as a unified reasoning-action sequence, effectively functioning as the ``brain'' of a generalist robot system.
\Cref{fig:demo-cases} illustrates \name{} in an interactive table-organization task, demonstrating flexible capabilities such as understanding complex instructions, handling real-time interruptions, monitoring task progress, and engaging in proactive dialogue to clarify ambiguous commands or infer user intent.

Modeling such complex interactive task execution within a single VLM is challenging. Although general VLMs have achieved strong performance in digital domains, extending them to physical robots is far more demanding: robots must continuously perceive and act in dynamic environments, interpret ambiguous instructions, adapt to real-time feedback, and make sequential decisions under strict physical and temporal constraints.
Addressing this gap requires overcoming two major limitations of existing models:
(1) limited embodied reasoning—the ability to ground objects and spatial concepts in the physical world and integrate these signals for adaptive planning and task-centric reasoning~\cite{team2025gemini};
(2) lack of flexible multimodal interaction—hindered both by its inherent complexity and by the scarcity of corresponding training data.

To address these challenges, \name{} is trained with a three-stage strategy:
\begin{itemize}[leftmargin=0.5cm]
    \item \textbf{Continued pretraining} on general VLMs to enhance foundational embodied reasoning capabilities. We curate a large-scale dataset covering various robot-relevant tasks, such as 3D spatial understanding, visual grounding, and task-centric reasoning, enabling the model to strengthen its grounded planning and reasoning abilities.
    \item \textbf{Supervised finetuning} to endow the model with complex interactive capabilities. We employ comprehensive data synthesis to incorporate chain-of-thought reasoning and model interactive task execution as a unified reasoning-action sequence. The synthetic data covers a full spectrum of capabilities, including complex instruction understanding, long-horizon planning, task status monitoring, dynamic replanning, real-time interruption handling, and human-robot dialogue.
    \item \textbf{Reinforcement learning} to further refine the reasoning ability and strengthen the consistency between reasoning and actions, particularly in long-horizon, interactive tasks.
\end{itemize}

We comprehensively evaluate \name{} on embodied reasoning and interactive task execution. Across 31 benchmarks covering robot-relevant abilities (3D spatial understanding, visual grounding, task-centric reasoning) and general-purpose skills (general VQA, multimodal reasoning), \name{} achieves obvious improvements on most robot-relevant tasks while maintaining strong general-purpose performance. On a curated interactive-task benchmark spanning in-distribution and out-of-distribution (OOD) settings and diverse instruction types (multi-stage, constrained, open-ended, invalid, interrupted), \name{} consistently outperforms commercial (\eg{}, GPT-4o, Gemini-2.5-Pro) and open-source (\eg{}, Qwen2.5-VL, RoboBrain-2.0) baselines; notably, \name{}-32B exceeds the strongest baseline, Gemini-2.5-Pro, by 3.0 and 11.8 percentage points in accuracy on the two OOD settings. We further assess five real-world scenarios—table bussing, grocery shopping, checkout packing, tableware organization \& shipment, and dietary filtering—using task-progress metrics in a hierarchical robot system under two low-level control modes (human teleoperation and an automatic VLA controller). Across both modes, \name{}-32B surpasses Gemini-2.5-Pro by 1.6 and 4.3 percentage points on task progress and markedly outperforms all other baselines by $28.1 \sim 64.6$ percentage points. Our experiments demonstrate that \name{} couples strong embodied reasoning with flexible high-level planning and interaction, advancing toward general-purpose embodied intelligence.

We summarize \name{}’s main features as follows:
\begin{itemize}[leftmargin=0.5cm]
    \item \textbf{Unified model}. \name{} is a single vision-language model that unifies robot reasoning, task planning, and human-robot interaction, enabling robots to follow complex instructions, plan long-horizon tasks, and interact naturally in an end-to-end manner.
    \item \textbf{Flexible interaction}. Within this unified framework, \name{} supports proactive dialogue to clarify ambiguity or infer user intent, real-time interruption handling that seamlessly incorporates feedback, and context-aware commonsense reasoning for complex, open-ended tasks.
    \item \textbf{Robust performance}. We assess \name{} in two setups: (i) on a curated interactive-task benchmark covering both in- and out-of-distribution scenarios with diverse instruction types, and (ii) across five real-world scenarios in a hierarchical robot system with both human teleoperation and an automatic VLA model as the low-level controller. These evaluations demonstrate that \name{} consistently delivers strong performance across all settings.
\end{itemize}

%% file: sections/2-model.tex
\section{The \name{} Model}
\label{sec:mira_model}

\begin{figure}
    \centering
    \includegraphics[width=0.95\linewidth]{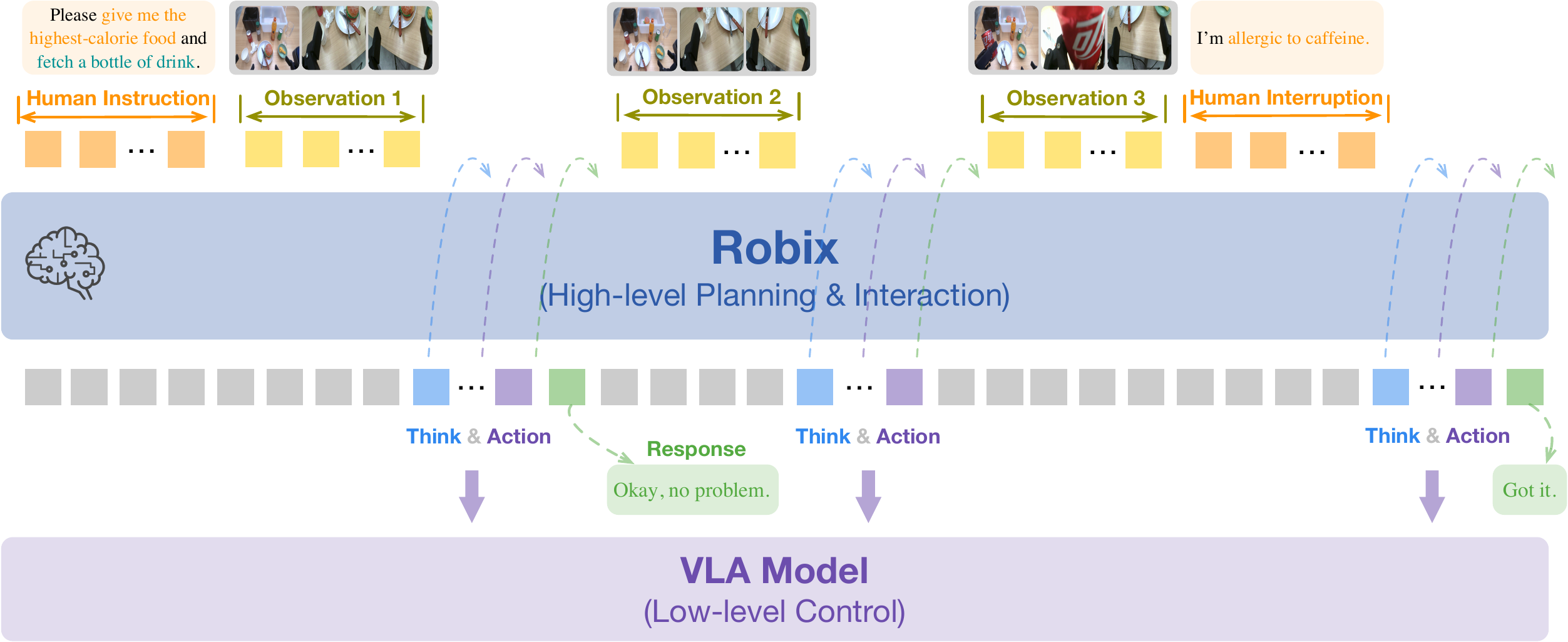}
    \caption{
        Illustration of the hierarchical robot system. \name{} serves as the high-level cognitive layer, interpreting tasks and reasoning over multimodal inputs to generate language responses and action plans. The low-level controller layer executes the corresponding atomic commands, enabling interaction with both humans and the physical environment.
    }
    \label{fig:mira-architecture}
\end{figure}

\Cref{fig:mira-architecture} illustrates the hierarchical robot system, where \name{} serves as the high-level cognitive layer responsible for planning and interaction. The low-level controller—typically implemented as a vision-language-action (VLA) model—executes the atomic commands generated by \name{}, enabling the robot system to directly interact with the physical environment.

At each iteration, \name{} directly processes visual observations from robot-mounted cameras and user utterances, selectively producing atomic action commands for the low-level controller and appropriate verbal responses. This iterative reasoning-action loop allows \name{} to perform deliberate reasoning and generate contextually grounded behaviors. The sequential decision-making process can be formally modeled as:
\begin{equation}
P\left(t_n, a_n, r_n ~\big|~ (o_1, u_1, t_1, a_1, r_1), \dots, [(o_{n-i}, u_{n-i}, t_{n-i}, a_{n-i}, r_{n-i})]_{i=1}^{N}, o_n, u_n\right).
\label{eq:formulation}
\end{equation}
Here, each step involves predicting the next thought $t_n$, action $a_n$, and optional verbal response $r_n$, conditioned on the current observations $o_n$, optional user instruction $u_n$ and the interaction history. These intermediate thoughts provide a structured reasoning trace that guides decision-making and enables nuanced, context-aware interaction with humans and the environment. To balance memory usage and maintain inference efficiency under token budget constraints (\eg{}, 32k context length), we retain only the latest $N$ visual observations as explicit input. The full sequence of prior thoughts and actions is stored in short-term memory, allowing \name{} to reason over recent history without exceeding capacity limits.

As shown in \Cref{fig:demo-cases}, \name{} unifies the entire interactive task execution process—including instruction understanding, task planning, task status monitoring, real-time user feedback integration, proactive dialogue, and dynamic replanning—through grounded, multi-faceted reasoning. Unlike prior modular frameworks for task planning or human-robot interaction, \name{} offers significantly greater flexibility, allowing the robot to adapt its behavior in real time to dynamic environmental changes, thereby achieving human-like adaptability.



%% file: sections/3-data_training.tex
\section{Training Recipe}
\label{sec:data_training}

We develop \name{}-7B and \name{}-32B by continually training Qwen2.5-VL-7B and 32B \cite{bai2025qwen2} on approximately 200 billion tokens using a three-stage training pipeline. First, we perform continued pretraining to enhance the model’s capabilities in robot-relevant perception and reasoning (\Cref{sec:ct}). Next, we apply supervised finetuning to model the complex human-robot interaction and long-horizon task planning as a sequential decision-making process grounded in chain-of-thought reasoning (\Cref{sec:sft}). Finally, we leverage reinforcement learning to further improve the embodied reasoning ability and enhance the alignment between reasoning and action in interactive long-horizon tasks (\Cref{sec:rl}).

\subsection{Continued Pretraining}
\label{sec:ct}

\begin{figure}
    \centering
    \includegraphics[width=\linewidth]{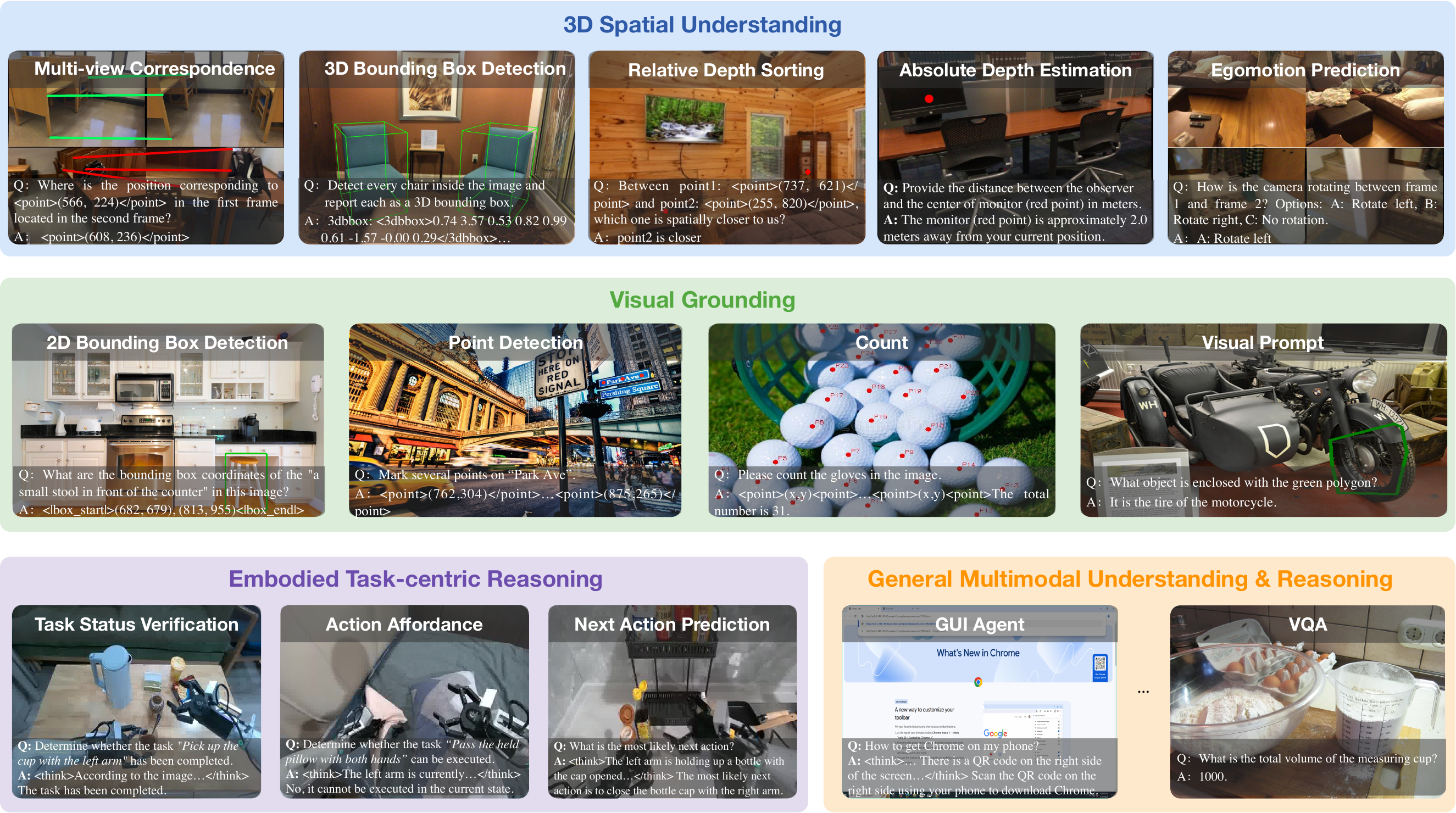}
    \caption{
        Overview of \name{}’s pretraining data, curated to strengthen core embodied reasoning capabilities (3D spatial understanding, visual grounding, and task-centric reasoning) while also supporting general multimodal understanding and reasoning. The figure showcases the diversity of the data, establishing a solid foundation for embodied models.
    }
    \label{fig:pretrain-data}
\end{figure}


A foundational capability of general-purpose embodied models is \textbf{embodied reasoning}—the ability to ground objects and spatial concepts in the physical world and integrate these signals for downstream robotic tasks~\cite{team2025gemini}. Our objective is to develop a vision-language model with embodied reasoning at its core, capable of generalizing across diverse embodied scenarios while maintaining strong foundational multimodal understanding. To support this, we construct a large-scale pretraining corpus comprising 200 billion high-quality and diverse tokens, targeting both robot-relevant and general-purpose multimodal capabilities, as illustrated in \Cref{fig:pretrain-data}. In particular, we emphasize 3D spatial understanding, visual grounding, and task-centric reasoning, while also incorporating general visual understanding, multimodal reasoning, and instruction tuning data. Below, we detail the data sources and task types used in our continued pretraining.

\paragraph{3D Spatial Understanding.}
Current VLMs generally lack strong spatial understanding capabilities, which is crucial for embodied scenarios such as navigation and manipulation planning.
To equip the model with 3D spatial understanding from 2D images, we curate over 30 million instruction pairs (about 40B tokens) spanning five key task types:
(1) Multi-view correspondence – learning 2D point correspondences across stereo or multi-view images of the same scene;
(2) 3D bounding box detection – predicting metric 3D bounding boxes from monocular images with open-vocabulary object descriptions;
(3) Relative depth sorting – inferring the depth ordering of objects within a single image;
(4) Absolute depth estimation – estimating absolute object depth using semantic masks and annotated depth maps;
(5) Egomotion prediction – modeling camera motion over time to support temporal and spatial reasoning.
The majority of the data is derived from the 3D spatial understanding training corpus of Seed-1.5-VL~\cite{guo2025seed1}. Additionally, part of the dataset is constructed from publicly available 3D benchmarks, including ScanNet\cite{dai2017scannet}, ScanNet++\cite{yeshwanth2023scannet++}, 3RScan\cite{wald2019rio}, CA-1M\cite{lazarow2025cubify}, SUN RGB-D\cite{song2015sun}, and ARKitScenes\cite{baruch2021arkitscenes}. Integrating these five spatial reasoning tasks effectively improves the model’s spatial awareness in embodied tasks.

\paragraph{Visual Grounding.} Visual grounding enables multimodal models to interpret user instructions and locate target objects in images. We use two grounding formats—bounding boxes and center points—and train on four types of data: 2D bounding box annotations, point annotations, counting, and visual prompt. We normalize all coordinate values to the range $[0, 1000]$, allowing consistent grounding predictions across varying image resolutions. 
The dataset comprises over 50 million instruction–response pairs (approximately 70 billion tokens), curated from both open-source resources and internal collections within the Seed-1.5-VL corpus. It encompasses a broad range of tasks, including:
(1) 2D bounding box annotations: predict bounding boxes from open-vocabulary descriptions, or generate textual descriptions given bounding box coordinates;
(2) Point annotations: predict object center points from descriptions or identify objects based on given coordinates.
(3) Counting: derived from bounding box and point data, supporting both box- and point-based counting via a two-stage localization and counting pipeline;
(4) Visual Prompt: prompts contain both textual instructions and visual annotations (\eg{}, points, bounding boxes, arrows), enabling the model to learn multimodal fusion and context-aware understanding grounded in visual cues.
Together, these tasks significantly enhance the model’s grounding abilities in both language-to-image and image-to-language directions, and improve its capacity for grounded planning in embodied settings.

\paragraph{Task-centric Reasoning.} To directly strengthen the model’s reasoning and planning abilities in embodied scenarios, we construct a large-scale embodied task-centric reasoning dataset based on publicly available robot and egocentric datasets, including AgiBot~\cite{bu2025agibot}, BridgeData V2~\cite{walke2023bridgedata}, Droid~\cite{khazatsky2024droid}, Egodex~\cite{hoque2025egodex}, RoboVQA~\cite{sermanet2024robovqa}, HoloAssist~\cite{wang2023holoassist}, and Ego4D~\cite{grauman2022ego4d}. We curate over 5 million examples (about 10B tokens) targeting three key reasoning functions:
(1) Task Status Verification---determining whether a task or subtask has been successfully completed;
(2) Action Affordance---assessing whether an action is feasible in the current context;
(3) Next Action Prediction---identifying the most plausible next step to achieve the intended goal.
To enrich the reasoning process, we further use Seed-1.5-VL-thinking~\cite{guo2025seed1} to generate step-by-step thought traces for our QA pairs via carefully designed prompts (details are shown in~\Cref{aapx:er}). This thought-augmented supervision enables the model to learn deliberate, high-level decision-making in dynamic and open-ended environments.

\paragraph{General Multimodal Reasoning.} To enhance the model’s general reasoning capabilities, we curate a diverse set of over 6 million multimodal instruction-image pairs (about 10B tokens) spanning STEM problem solving, agent-based decision making, and visual inference tasks. Specifically, we include:
(1) STEM Reasoning Data: Multimodal problem-solving examples in mathematics, physics, chemistry, and biology, combining textual questions with diagrams, equations, and visual content.
(2) Multimodal Agent Data: GUI-based agent demonstrations that involve step-by-step planning, error correction, and reflective reasoning. 
(3) Visual Inference Data: Tasks that require grounded visual reasoning, including spotting differences between paired images and generating HTML/CSS code from user interface screenshots.
Together, these datasets equip the model with robust abstract reasoning and cross-modal problem-solving abilities, supporting its generalization to complex tasks in open environments.

\paragraph{General Multimodal Understanding.} To preserve and enhance broad vision-language understanding, we curate a large-scale dataset of over 50 million image–text pairs (over 80B tokens) that serve as the foundation for multimodal comprehension.
(1) VQA: A diverse set of image- and video-based question answering tasks covering visual perception, factual knowledge, grounding, temporal reasoning, spatial understanding, and counting.
(2) Captioning: Dense captions for both images and videos, supporting the model’s understanding of static scenes and multi-frame temporal dynamics.
(3) OCR: To improve text recognition, we include large-scale annotated and synthetic datasets covering scene text, documents, tables, charts, and flowcharts.
The dataset is filtered from both open-source resources and the Seed-1.5-VL corpus, together forming a robust foundation for training general-purpose vision–language models.

\paragraph{Instruction Tuning.}
To further enhance the model’s instruction-following and reasoning abilities, we construct a high-quality instruction tuning dataset comprising 1 million examples. These examples span a wide range of tasks and are built by extracting curated subsets from previously collected data, integrating both general instructions and chain-of-thought examples from open-source and internal sources. We refine the instructions using Seed-1.5-VL~\cite{guo2025seed1} for quality filtering and ensure better alignment between instructions, images, and responses. This instruction-tuned dataset significantly improves the model’s ability to follow open-world multimodal instructions and engage in multi-turn, grounded reasoning.

We adopt a two-stage training strategy leveraging the large-scale, diverse corpus described above. In stage 1, we continue pretraining Qwen2.5-VL~\cite{bai2025qwen2} on the full dataset—comprising approximately 5\% text-only data—updating all model parameters to enhance general multimodal and embodied reasoning capabilities. Training follows a full cosine learning rate schedule, starting at $1\times10^{-5}$ and decaying to $1\times10^{-6}$, with linear warm-up over the first 10\% of total steps. We use a sequence length of 32,768 tokens, with effective batch sizes of 1536× and 3008× the sequence length for the 7B and 32B models, respectively.
In stage 2, we perform instruction tuning on curated instruction-following data to align the model with multimodal prompts and improve instruction adherence. The vision encoder is frozen during this phase, while all other parameters remain trainable. The learning rate is fixed to the final value from Stage 1 ($1\times10^{-6}$) and remains constant throughout Stage 2. Optimizer states are carried over from Stage 1, and no additional warm-up is used.
Both stages are optimized using AdamW~\cite{2015-kingma,loshchilov2018decoupled}, with $\beta_1 = 0.9$, $\beta_2 = 0.99$, and a weight decay of 0.01.
Training on this diverse and comprehensive corpus significantly improves the model’s embodied reasoning, multimodal understanding, and its ability to generalize to long-horizon, interactive tasks in real-world settings.



\subsection{Supervised Finetuning}
\label{sec:sft}



    

\begin{figure}[t]
    \centering
    \includegraphics[width=1.0\linewidth]{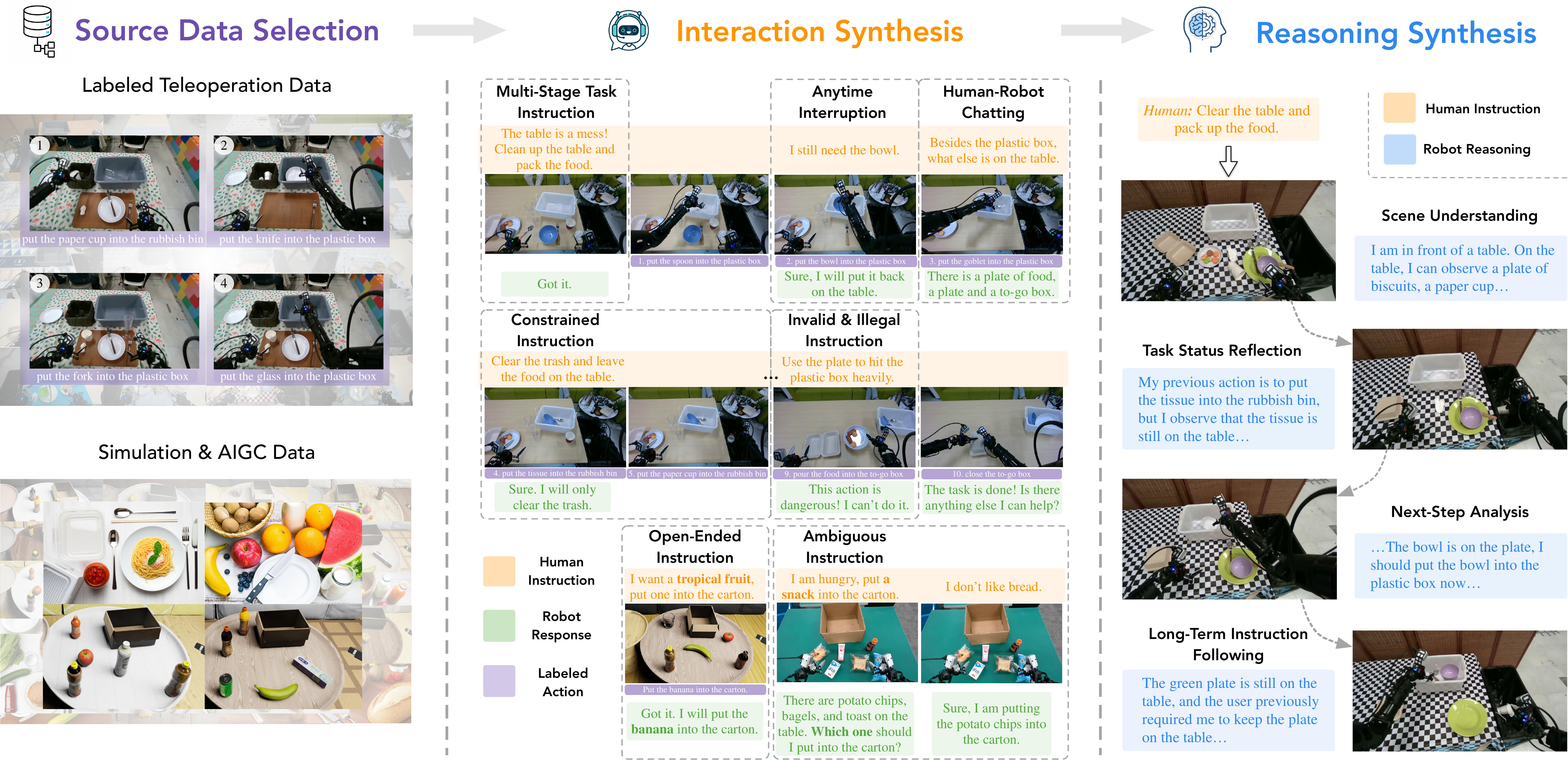}
    \caption{
        Our data synthesis pipeline. The leftmost panel shows the data sources; the center illustrates how diverse human-robot interactions are synthesized from these sources; the rightmost panel presents example snapshots of the generated reasoning traces.
    }
    \label{fig:sft-data-pipeline}
\end{figure}

The supervised fine-tuning (SFT) stage adapts the preceding pretrained model into the robot’s high-level cognitive module while retaining its original capabilities. A central challenge lies in the scarcity of large-scale, multi-turn egocentric-vision datasets that integrate human-robot interaction with task planning. To address this, we design a data-synthesis pipeline that transforms existing task-planning datasets into human–robot interaction trajectories. Two properties of the resulting SFT data are crucial for out-of-distribution generalization: (1) \textbf{diverse human–robot interactions} and (2) \textbf{high-quality reasoning traces}. The overall pipeline is shown in \Cref{fig:sft-data-pipeline}; the interaction and reasoning synthesis modules are detailed below.

\subsubsection{Interaction Synthesis}

The interaction synthesis is mainly based on two data sources:
\begin{itemize}[leftmargin=0.5cm]
    \item Teleoperated Robot Demonstrations. We utilize both internal teleoperation data (previously used to train GR-3~\cite{cheang2025gr}) and the open-source AGIBot dataset~\cite{bu2025agibot}. Each contains episodes of robots performing diverse tasks (\eg{}, table bussing, breakfast preparation). Human annotators segment each demonstration into clips, where each clip corresponds to an atomic action—\eg{}, ``put the tissue into the trash bin''.
    \item Simulation \& AIGC Data. We programmatically generate diverse object organization scenarios in our in-house simulator, and further employ state-of-the-art text-to-image models~\cite{gong2025seedream20nativechineseenglish} to synthesize complex scenes with items not yet supported in the simulator. To ensure quality, we apply both model-based and human-in-the-loop filtering to discard low-quality generations.

\end{itemize}

Building on the above sources, we define seven categories of human–robot interaction instructions and develop tailored data-synthesis strategies for each. Training on this dataset equips \name{} with flexible interaction capabilities, such as complex instruction understanding, real-time interruption handling, and proactive dialogue.

\paragraph{Multi-Stage Instruction.} Teleoperated trajectories are annotated with a task name (\eg{}, ``clean up the table and pack the food''). We select trajectories containing at least ten atomic actions and synthesize a corresponding user instruction from the task name, \eg{}, ``\textit{The dining table is a mess. Please clean the table and pack the food on the plate}.''

\paragraph{Constrained Instruction.} We partition each teleoperation trajectory into non-overlapping segments where feasible. For example, a table-bussing task can be decomposed into subtasks such as garbage collection, tableware collection, and food packing. Based on these subtasks, we synthesize tailored user instructions, \eg{}, ``\textit{Clean up the table while leaving the food on the table}'' and ``\textit{Remove the trash without moving other items}''.

\paragraph{Open-Ended Instruction.} We generate random scenes in simulation and prompt state-of-the-art LLMs to produce open-ended, commonsense instructions conditioned on each scene (\eg{}, ``\textit{Place the drink with the least sugar into the carton}'' for a scene containing Sprite, Coke, orange juice, and soda water). To include items currently not supported by the simulator (\eg{}, hamburgers, spaghetti, watermelons), we pair such instructions with images synthesized by advanced text-to-image models. Because current text-to-image models still struggle with instruction following and image quality, we apply automated and human-in-the-loop filtering, remaining only 10\% of the original dataset after filtering.

\paragraph{Anytime Interruption.}
We curate a variety of user interruption utterances (\eg{}, ``\textit{Stop!}'', ``\textit{Hold on. I still need it}'', ``\textit{Wait, put the fork into the sink first}'') and randomly inject them into task flows. Corresponding robot responses are then synthesized using timing-aware heuristics: if the interruption occurs before grasping, the robot halts or adjusts its plan; if it occurs after grasping, the robot returns the item to the table and replans. Such anytime interruption handling is critical for long-horizon tasks, as mid-task feedback and error correction substantially enhance system robustness.

\paragraph{Invalid Instruction.} To mitigate hallucination and prevent robots from engaging in dangerous actions, we synthesize four types of invalid instructions as follows:
(1) Instructions asking the robot to manipulate items that do not exist in the scenario;
(2) Instructions requiring physically impossible actions, such as ``\textit{Put the table into the rubbish bin}'';
(3) Instructions demanding abilities beyond the robot’s current capabilities, \eg{}, ``\textit{Open the coke for me}'';
(4) Unsafe or dangerous commands, \eg{}, ``\textit{Throw the knife onto the sofa}''.
For these invalid or illegal instructions, we design corresponding response strategies to enable the robot to refuse compliance with the user's requests.

\paragraph{Ambiguous Instruction.} To enable our model to clarify ambiguous instructions, we construct scenes with multiple similar items (\eg{}, apple, orange, pear) and synthesize underspecified instructions(\eg{}, ``\textit{Put a fruit into the basket}''). Training with these data enable the model to seek clarification when needed—a capability essential for robust robot systems.

\paragraph{Chat Instruction.} We develop some heuristics to randomly insert short human-robot dialogue segments at context-appropriate times. For example, when the robot is collecting trash from the table, the user may ask ``\textit{I want some fruit. What kind of fruit is on the table?}''. This type of instruction requires the robot to respond verbally rather than performing any physical manipulation.

\subsubsection{Reasoning Synthesis}

To incorporate chain-of-thought reasoning, we prompt state-of-the-art VLMs to generate high-quality reasoning traces emphasizing (1) scene understanding, (2) task status reflection, (3) long-term instruction following, and (4) next-step analysis.
\begin{itemize}[leftmargin=0.5cm]
    \item \textbf{Scene understanding.} This part of reasoning enables the robot to accurately identify task-relevant, operable objects in the current scene, with emphasis on those within the robot’s field of view.
    \item \textbf{Task status reflection.} Robots should be capable of reflecting on their prior actions and repeating tasks when initial attempts fail. Furthermore, they need to identify key milestones in long-horizon tasks and proactively request human assistance upon encountering irrecoverable errors. This capability is also critical for handling user interruptions, as robots must maintain awareness of their current status to plan subsequent actions (\eg{}, tracking whether a gripper is holding an item).
    \item \textbf{Long-term instruction following.} This module is designed to help robot persist the initial goal and intermediate user instructions across long-horizon tasks, ensuring the primary objective is completed and mid-task instructions continue to guide actions many steps later (\eg{}, ``\textit{After cleaning the table, grab me a drink from the fridge}'').
    \item \textbf{Next-step analysis.} In the final phase of reasoning, the robot should analyze potential actions for the next step when the overall task remains incomplete. This analysis include assessing target reachability and whether executing the action advances overall task completion.
\end{itemize}

Inspired by UI-TARS \citep{qin2025ui}, we adopt \textit{ActRe}~\citep{yang2024react} and \textit{Thought Bootstrapping}~\citep{qin2025ui} to synthesize high-quality reasoning traces. Unlike conventional LLM reasoning, robot reasoning must remain concise to support real-time interaction. Accordingly, we prompt Seed-1.5-VL to produce succinct traces (within 200 tokens). We also apply a model-based filtering pipeline to discard hallucinated or logically inconsistent reasoning. These high-quality, multi-faceted chain-of-thought traces enable \name{} to execute robust long-horizon task planning with task-status monitoring and dynamic replanning.




\subsection{Reinforcement Learning}
\label{sec:rl}

Following the supervised fine-tuning (SFT) stage, the model exhibits promising agentic capabilities in adaptive task planning and natural human-robot interaction. However, several limitations in robot reasoning and planning persist, notably: (1) \textit{irrational reasoning}, such as generating conflicting thoughts, lacking common sense, or partially disregarding user instructions; and (2) \textit{thought-action inconsistency}, where the model’s proposed plan diverges from its preceding thought in intent or content. For example, in a table-cleaning task, the SFT model correctly infers that a tissue left on the table should be discarded in a rubbish bin. Yet, in the subsequent plan, it incorrectly suggests handling a paper cup instead. These issues negatively impact the model's effectiveness in real-world task execution.

To mitigate these problems, we adopt reinforcement learning (RL), specifically Group Relative Policy Optimization (GRPO) \citep{shao2024deepseekmath, guo2025deepseek}, to enhance both the reasoning capacity and the coherence between thought and action. Our approach is based on two core strategies: (1) \textbf{co-training with general visual reasoning data}, and (2) \textbf{reward design targeting thought-action consistency}.

\paragraph{Co-Training with General Visual Reasoning Data.} The RL stage utilizes two primary data sources: robot interaction data and general visual reasoning datasets. Training on robot interaction data improves the model's robustness and generalization to out-of-distribution (OOD) scenarios. Meanwhile, incorporating general visual reasoning data strengthens the model's inherent reasoning capabilities. This co-training strategy helps alleviate irrational reasoning and enhances overall task understanding and solving. The general visual reasoning datasets include a wide range of cognitive challenges, such as task completion verification, action affordance evaluation, and object localization—covering a broad spectrum of reasoning skills relevant to real-world robot applications.

\paragraph{Reward Design for Thought-Action Consistency.} To explicitly encourage alignment between the model's thought and action, we introduce a thought-action consistency reward in addition to standard rewards for output formatting and action accuracy. At each decision step, the model's generated thought and corresponding action are extracted and evaluated by an external LLM (Qwen-2.5-32B \citep{yang2024qwen2} in our experiments). This auxiliary reward model is prompted to assess whether the action is logically consistent with the preceding thought. A negative reward is given if the assessment indicates inconsistency. The system prompt of the reward model is listed in \Cref{appendix:prompts}.

To maximize the effectiveness of RL training, we also employ a data filtering procedure designed to retain only samples that can provide meaningful gradient information for GRPO. The key idea is to discard questions whose candidate answers exhibit low reward variance, as such samples contribute little to policy improvement. Specifically, for each question in the dataset, we generate multiple candidate answers using the SFT model and remove those with low variance in their rewards:
\begin{equation}
    \mathcal{D}_{\text{new}} = \left\{ (x_n, y_n^*) \in \mathcal{D} \;\middle|\; \operatorname{Var}\left( \left\{ R(y_n^{(i)}, y_n^*) \right\}_{i=1}^M \right) > \tau,\; y_n^{(i)} \sim \pi_{\text{SFT}}(\cdot \mid x_n) \right\},
\end{equation}\label{eq:rl-filter}
where $\mathcal{D}$ denotes the original dataset, $R(y_n^{(i)},y_n^*)$ is the reward function assigning a scalar score to the $i$-th generated answer $y_n^{(i)}$ based on the ground-truth $y_n^*$ for question $x_n$, and $\pi_{\text{SFT}}$ is the base policy for RL. The definitions of input $x_n$ and output $y_n^*$ follow \Cref{eq:formulation}: $x_n$ consists of the current observation, instruction, and trajectory, while $y_n$ comprises the model’s thought, an optional action, and an optional robot response. In our experiments, we set the number of samples $M$ to $8$ and the variance threshold $\tau$ to $0$. All RL training is performed using the verl framework~\citep{sheng2024hybridflow}.

Through the combination of co-training with diverse reasoning data and targeted reward design, our reinforcement learning strategy substantially improves the model's generalization to novel tasks and enhances the consistency between reasoning and planning.

%% file: sections/4-experiments.tex
\section{Experiments}
\label{sec:experiments}

We conduct extensive experiments to comprehensively evaluate the performance of \name{}, focusing on the following key questions:
\begin{itemize}[leftmargin=0.5cm]
\item Does \name{} enhance fundamental embodied reasoning capabilities?
\item Can \name{} effectively model the full process of interactive task execution in an end-to-end manner? 
\item Does \name{} generalize well to out-of-distribution tasks?
\item How does the full robotic system perform when integrating \name{} with a VLA model on real-world tasks?
\end{itemize}

We first evaluate fundamental embodied reasoning capabilities of our model and baseline methods on public benchmarks (\Cref{sec:pretrain_exp}). 
Next, we assess the planning and interaction abilities of our model and baseline methods with both offline, pre-defined test sets (\Cref{sec:offline_exp}) and online robotic tasks (\Cref{sec:online_exp}). 

\subsection{Fundamental Perception \& Reasoning Evaluation}
\label{sec:pretrain_exp}

We evaluate \name{} after continued pretraining (denoted as \name{}-Base) on a comprehensive set of public benchmarks against state-of-the-art multimodal models including Qwen-2.5-VL-7B\&32B~\cite{bai2025qwen2}, RoboBrain-2.0-32B~\cite{team2025robobrain}, Cosmos-Reason1-7B~\cite{azzolini2025cosmos}, Gemini-2.5-Pro~\cite{team2025gemini}, OpenAI GPT-4o~\cite{hurst2024gpt}, Seed-1.5-VL and Seed-1.5-VL-Think~\cite{guo2025seed1}. The evaluation spans (1) robotics-relevant embodied reasoning (3D spatial understanding, visual grounding, task-centric reasoning) and (2) general multimodal understanding and reasoning. \Cref{tab:pretrain_eval} presents the detailed results. 

\begin{table}[t]
\begin{threeparttable}
\resizebox{\textwidth}{!}{
\begin{tabular}{lcccccc|cccc}
\toprule
\textbf{Benchmark} & \textbf{\begin{tabular}[c]{@{}c@{}}\name{}\\ 7B Base\end{tabular}} & \textbf{\begin{tabular}[c]{@{}c@{}}\name{}\\ 32B Base\end{tabular}} & \textbf{\begin{tabular}[c]{@{}c@{}}Qwen 2.5\\ VL 7B\end{tabular}} & \textbf{\begin{tabular}[c]{@{}c@{}}Qwen 2.5\\ VL 32B\end{tabular}} & \textbf{\begin{tabular}[c]{@{}c@{}}Cosmos\\ Reason1 7B\end{tabular}} & \textbf{\begin{tabular}[c]{@{}c@{}}RoboBrain\\ 2.0 32B\end{tabular}} & \textbf{\begin{tabular}[c]{@{}c@{}}Gemini\\ 2.5 Pro\end{tabular}} & \textbf{\begin{tabular}[c]{@{}c@{}}OpenAI\\ GPT 4o\end{tabular}} & \textbf{\begin{tabular}[c]{@{}c@{}}Seed\\ 1.5-VL\end{tabular}} & \textbf{\begin{tabular}[c]{@{}c@{}}Seed\\ 1.5-VL Think\end{tabular}} \\
\midrule
\multicolumn{11}{c}{\textbf{3D Spatial Understanding}}   \\ \hline
VSIBench   & 44.6   & \textbf{50.9}   & 38.2   & 39.1 &  33.9  & 45.2 & \textbf{43.4}$^{*}$  & 42.5$^{*}$ & 34.0$^{*}$ & 39.5$^{*}$ \\
BLINK & \textbf{87.6}  & 83.5 & 86.5   & 78.3 &  69.7 & 82.4 & \textbf{87.6}$^{*}$  & 80.5$^{*}$ & 79.0$^{*}$ & 82.0$^{*}$\\
CV-Bench   & 86.5   & \textbf{87.4}   & 79.0  & 81.6 & 76.0  & 83.7 &  \textbf{85.7}$^{*}$  & 79.7$^{*}$  & 77.9$^{*}$ & 82.4$^{*}$\\
EmbSpatial & 77.4 & \textbf{79.0}   & 70.1   & 72.3 & 64.6 & 76.6 & \textbf{77.2}$^{*}$ & 70.6$^{*}$ & 68.0$^{*}$ & 73.5$^{*}$ \\
SAT& 71.1 & 79.6  & 52.8 & 74.9 & 60.7  & \textbf{80.3} & \textbf{79.5}$^{*}$  & 62.4$^{*}$ & 67.5$^{*}$ & 72.7$^{*}$\\
VSR& 83.3 & 83.7   & 82.2  & 83.6 & 79.9  & \textbf{84.0} & \textbf{82.5}$^{*}$  & 77.5$^{*}$ & 78.1$^{*}$ & 80.1$^{*}$\\
SpatialBench  & 64.7 & 65.4   & 61.7   & 66.8 & 61.6 & \textbf{68.8} & \textbf{66.0}$^{*}$  & 63.7$^{*}$ & 64.0$^{*}$ & 66.3$^{*}$ \\
DA-2k & 72.2 & \textbf{77.1}   & 65.0 & 68.6 & 65.8  & 56.9 & 83.0$^{*}$ & 78.2$^{*}$ & 86.5$^{*}$ & \textbf{87.5}$^{*}$ \\
\hline
\multicolumn{11}{c}{\textbf{Visual Grounding}}\\ \hline
LVIS-MG    & 70.2    & \textbf{79.2}    & 30.6  & 54.2 & 18.2 & -$^{\dagger}$  & 63.8 & -$^{\dagger}$  & \textbf{73.8}  & 72.5 \\
Refcoco$_{val}$& 89.2 & \textbf{91.5}   & 88.9  & 89.3   & 77.1   & -$^{\dagger}$  & 74.6   & -$^{\dagger}$ & \textbf{92.8}$^{*}$ & 92.6$^{*}$\\
Refcoco+$_{val}$& 83.8 & \textbf{86.2}  & 82.6  &  83.2  &  69.2  &  -$^{\dagger}$  &  -$^{\dagger}$  & -$^{\dagger}$ & 89.3$^{*}$ & \textbf{89.7}$^{*}$\\
Refcocog$_{val}$& 87.1 &  \textbf{89.0}  &  85.6  &  87.3  &  75.0  &  -$^{\dagger}$   &  -$^{\dagger}$  & -$^{\dagger}$ & 90.1$^{*}$ & \textbf{90.4}$^{*}$ \\
Refcocou$_{val}$& 87.1 &  \textbf{88.6}  &  85.5  &  86.9  &  74.2  &  -$^{\dagger}$   &  -$^{\dagger}$  & -$^{\dagger}$ & \textbf{88.9}$^{*}$ & 88.5$^{*}$ \\
VisualWebBench& 63.2 & \textbf{68.9}   & 59.4 & 66.1 & 51.5 & 20.5  & 87.3  & 80.2 & \textbf{88.0} & 87.3\\
Pixmo-Point & 29.5 & \textbf{47.3} & 43.5 & 41.3 & 6.0 & 46.0 & 11.3$^{*}$ & 10.8$^{*}$ & \textbf{13.7}$^{*}$ & 9.8$^{*}$ \\
Where2Place & 41.9 & 45.2 & 33.0 & 39.9 & 11.4 & \textbf{73.6} & \textbf{39.9}$^{*}$ & 26.9$^{*}$ & 18.4$^{*}$ & 25.4$^{*}$\\
\hline
\multicolumn{11}{c}{\textbf{Embodied Task-centric Reasoning}} \\ \hline
Agibot-ER  & 61.0    & \textbf{62.6}   & 48.2 & 55.4 & 38.0  & 54.3   & \textbf{67.1}$^{*}$  & 49.9$^{*}$ & 63.6$^{*}$ & 60.5$^{*}$\\ 
EgoTaskQA  & 28.6   & \textbf{33.6}  & 23.7 & 31.3 & 28.4 & 31.5  & \textbf{37.0}$^{*}$  & 30.0$^{*}$ & 27.8$^{*}$ & 31.1$^{*}$\\
OpenEQA-hm3d  & 48.8  & \textbf{51.1}  & 42.2  & 46.8  & 40.0  & 48.7  & \textbf{63.8}$^{*}$ & 63.4$^{*}$ & 58.8$^{*}$ & 60.8$^{*}$\\
OpenEQA-scannet  &  58.0   & \textbf{58.9}   & 51.3 & 52.8  & 50.9  & 55.9  & \textbf{74.3}$^{*}$ & 71.3$^{*}$ & 66.1$^{*}$ & 68.3$^{*}$\\
ERQA  & 42.5 & 43.5  & 38.0 & 42.8 & 39.3 & \textbf{46.0}  & \textbf{55.0}$^{*}$  & 47.8$^{*}$ & 39.8$^{*}$ & 47.5$^{*}$\\
RoboVQA   & 53.6    & 48.3   & 54.0 & \textbf{60.0}  & 55.4  & 55.7   & 33.9$^{*}$  & 34.5$^{*}$ & \textbf{37.0}$^{*}$ & 35.1$^{*}$\\
\hline
\multicolumn{11}{c}{\textbf{Multimodal Understanding \& Reasoning}}\\ \hline
MME& 2332.8  & 2427.2   & 2273.1  & 2425.5 & 2150.0 & \textbf{2462.3} & \textbf{2491.3}$^{*}$  &  2271.9$^{*}$ & 2314.4$^{*}$ & 2470.8$^{*}$\\
MMbench    & 87.6 & 89.1  & 86.8 & \textbf{89.2} & 85.1  & 88.6  & \textbf{90.1}  & 84.3 & 88.0 & 89.9\\
RealworldQA& \textbf{70.7} & 69.0 & 65.9 & 68.4 & 67.8  & 63.9   & 78.0  & 76.2 & 77.0 & \textbf{78.4} \\
SimpleVQA  & 44.7 & \textbf{49.0}   & 47.7 & 45.2   & 42.5  & 41.9   & 62.0  & 50.1 & 63.1 & \textbf{63.4}\\
EgoSchema  & 66.2   & 73.4  & 67.6 & \textbf{74.0} & 59.8  & 70.8 & 74.1$^{*}$ &  69.8$^{*}$ & 64.3$^{*}$ & \textbf{77.2}$^{*}$ \\
VideoMME   & 63.7 & 67.6  & 65.3 & \textbf{68.8} & 61.9 & 66.9  &  \textbf{86.9} & 71.9 & 77.6 & 77.9\\
NextQA& 82.5 & 81.6  & \textbf{82.9} & 81.8 & 79.5  & 78.7  & \textbf{83.5}$^{*}$  & 72.7$^{*}$ &  62.6$^{*}$ & 68.6$^{*}$ \\ 
MathVista  & 68.0 & 69.6 & 68.3 & \textbf{75.6} & 62.1 & 68.7 & 82.7 & 63.8 & 83.0 & \textbf{85.6} \\
MMMU  & 51.6 & 58.9 & 51.9 & 57.3 & 47.0  & \textbf{59.2}  & \textbf{81.7} & 70.7 & 73.6 & 77.9\\ \bottomrule
\end{tabular}
}
\begin{tablenotes}
\footnotesize
\item[$^{*}$] Collected by ourselves via API in July 2025.
\item[$^{\dagger}$] Invalid results due to failures in following format requirements.
\end{tablenotes}
\end{threeparttable}
\caption{Performance of \name{} on public vision-language benchmarks compared to prior models. The left side shows \name{} and state-of-the-art open-source baselines, while the right side presents closed-source large commercial models. The highest score in each benchmark is highlighted in bold within each group.}
\label{tab:pretrain_eval}
\end{table}

\paragraph{3D Spatial Understanding.}
We evaluate \name{}’s 3D spatial understanding across 8 spatial reasoning benchmarks: VSIBench~\cite{yang2025thinking}, BLINK~\cite{fu2024blink}, CV-Bench~\cite{tong2024cambrian}, EmbSpatial~\cite{du2024embspatial}, SAT~\cite{ray2024sat}, VSR~\cite{liu2023vsr}, SpatialBench~\cite{du2024embspatial}, and DA-2k~\cite{yang2024depth}. Details of these benchmarks are provided in~\Cref{appx:base}. As shown in \Cref{tab:pretrain_eval}, \name{}-7B and \name{}-32B outperform their backbones (Qwen2.5-VL-7B/32B) on 7 of 8 spatial reasoning tasks, with average accuracies of 73.4 and 75.8 compared to 66.9 and 70.7, respectively. They also surpass embodied models Cosmos-Reason1-7B (64.0) and RoboBrain-32B (72.2), and exceed the best commercial baseline, Gemini-2.5-Pro, on 5 of 8 tasks.

\paragraph{Visual Grounding.} We evaluate \name{}’s visual grounding capabilities on eight benchmarks covering both bounding-box and center-point tasks, including LVIS-MG~\cite{gupta2019lvis}, RefCOCO~\cite{yu2016modeling}, VisualWebBench~\cite{liu2024visualwebbench}, Pixmo-Point~\cite{deitke2024molmo}, and Where2Place~\cite{yuan2024robopoint}. \name{} consistently outperforms its backbone across all benchmarks and surpasses state-of-the-art commercial models on most tasks. Notably, \name{}-7B and \name{}-32B improve the absolute F1 score on LVIS-MG by 39.6 and 25.0 points over Qwen2.5-VL-7B and 32B, respectively. \name{}-32B also outperforms commercial models on most tasks. These results highlight \name{}’s strong performance in object localization, pointing and fine-grained visual understanding.

\paragraph{Task-centric Reasoning.} Embodied reasoning reflects a model’s ability to understand and reason about robotic tasks. We evaluate \name{} across 5 diverse open benchmarks, including ERQA~\cite{kirillova2021erqa}, RoboVQA~\cite{sermanet2024robovqa}, OpenEQA (HM3D \& ScanNet)~\cite{majumdar2024openeqa}, and EgoTaskQA~\cite{jia2022egotaskqa}. In addition, we introduce Agibot-ER, a real-world task reasoning benchmark derived from the Agibot dataset~\cite{bu2025agibot}, which includes manually annotated test sets of 97, 120, and 381 samples for the three key reasoning tasks—Task Status Verification, Action Affordance, and Next Action Prediction, respectively. Full details of the benchmark are provided in~\Cref{aapx:er}. We report the average results of the three tasks on this benchmark. \name{} consistently outperforms its backbone models as well as Cosmos-Reason1-7B and RoboBrain-2.0-32B across most benchmarks. On Agibot-ER, \name{} delivers substantial gains over its backbones, improving absolute accuracy by 12.8 and 7.2 points for the 7B and 32B versions, respectively. It further surpasses Cosmos-Reason1-7B and RoboBrain-2.0-32B by 23 and 8.3 points, demonstrating superior performance in embodied, task-centric reasoning.

\paragraph{General Multimodal Understanding \& Reasoning} 
Multimodal understanding is a core capability of vision–language models and a primary focus of VLM development. To evaluate both static image and dynamic video understanding, we assess \name{} on a suite of general VQA benchmarks—image-based (MME~\cite{fu2024mme}, MMBench~\cite{liu2024mmbench}, RealWorldQA~\cite{realworldqa}, SimpleVQA~\cite{cheng2025simplevqa}) and video-based (EgoSchema~\cite{mangalam2023egoschema}, VideoMME~\cite{fu2025videomme}, NextQA~\cite{xiao2021nextqa}). We further test general reasoning on MathVista~\cite{lu2023mathvista} and MMMU~\cite{yue2024mmmu}, which cover complex mathematical and multimodal problem-solving tasks. \name{} preserves the performance of its backbone on most benchmarks, demonstrating the benefit of training with diverse, high-coverage multimodal data, but still trails large-scale commercial models—underscoring the need to scale both data and model size for stronger general-purpose multimodal reasoning.

Overall, \name{} greatly enhances robotics-relevant perception and reasoning—particularly in 3D spatial understanding and visual grounding—while maintaining strong performance on general multimodal tasks. These gains deepen its understanding of spatial and temporal properties, enabling more effective reasoning and planning in real-world environments.

\subsection{Offline Evaluation}
\label{sec:offline_exp}

The offline evaluation enables fully automated assessment of planning and interaction capabilities using predefined evaluation sets. To thoroughly evaluate both interactive long-horizon planning and out-of-distribution (OOD) generalization, we design three dedicated evaluation sets:
\begin{itemize}[leftmargin=0.5cm]
    \item AGIBot Evaluation Set. We manually select 16 high-frequency daily tasks from the AGIBot dataset (\eg{}, making a sandwich, washing dishes with a dishwasher, arranging a sofa, washing clothes with a washing machine, arranging flowers) and ensure none appear in the training data. This set primarily evaluates the model’s long-horizon task planning capability on OOD tasks. Details are provided in~\Cref{appx:agibot}.
    \item Internal Out-of-Distribution (OOD) Benchmark. We manually design 16 scripts covering task planning and diverse human–robot interaction scenarios, including table organization, dietary filtering, checkout packing, grocery shopping, and shoe cabinet organization. These scripts are enacted by human participants—one acting as the user and the other executing actions via robot teleoperation or a Universal Manipulation Interface (UMI)~\cite{chi2024universal} device—and subsequently annotated by trained annotators. The benchmark includes tasks and items absent from the training data and is intended to evaluate interactive task execution in unseen scenarios.
    \item Internal In-Distribution (ID) Benchmark. This evaluation set is randomly sampled from our synthesized data and categorized by task type and user instruction into six groups: (1) multi-stage instructions, (2) constrained instructions, (3) invalid instructions, (4) user interruptions, (5) fail-and-replan, and (6) open-ended instructions. Each category targets evaluation of the model’s corresponding instruction following and task planning capabilities.
\end{itemize}

\paragraph{Data Format.} The overall statistics of each evaluation set are shown in \Cref{tab:eval_statistics}. Each episode is structured as a multi-turn dialogue and evaluated using a teacher-forcing approach, \ie{}, the model observes an error-free interaction and planning history when predicting the next step action. Because an observation–instruction pair may permit multiple valid next actions, we annotate a candidate action list for each step to capture all acceptable options. Examples of the offline evaluation format are provided in \Cref{appendix:offline_example}.

\begin{table}[t]
\small
  \centering
  \begin{tabular}{@{}m{1.5cm}m{1cm}m{1.5cm}cccccc}
    \toprule
     & \multirow{2}{*}{\textbf{AGIBot}} & \multirow{2}{*}{\textbf{\begin{tabular}[c]{@{}c@{}}Internal\\ OOD\end{tabular}}} & \multicolumn{6}{c}{\textbf{Internal ID}} \\ 
     &  &  & Multi-Stage & Constrained & Interrupt & Open-Ended & Invalid & Replan  \\ \hline\noalign{\vskip 3pt} 
    \# episodes & 16 & 16 & 9 & 25 & -- & 15 & -- & -- \\
    \# data & 142 & 225 & 119 & 233 & 110 & 15 & 60 & 100 \\
    \bottomrule
  \end{tabular}
  \caption{Statistics of the offline evaluation sets.}
  \label{tab:eval_statistics}
\end{table}

\begin{table}[t]
\small
  \centering
  \begin{tabular}{@{}m{3cm}m{1.5cm}m{1.5cm}cccccc}
    \toprule
     & \multirow{2}{*}{\textbf{AGIBot}} & \multirow{2}{*}{\textbf{\begin{tabular}[c]{@{}c@{}}Internal\\ OOD\end{tabular}}} & \multicolumn{6}{c}{\textbf{Internal ID}} \\ 
     &  &  & Multi. & Const. & Interrupt & Open. & Invalid & Replan  \\ \hline\noalign{\vskip 3pt} 
       & \multicolumn{6}{c}{\textit{Plan Accuracy}} & \textit{F1 score} & \textit{F1 score}\\ \hline\noalign{\vskip 3pt} 
     {\centering Gemini-2.5-Pro} & \underline{52.6}\ & \underline{83.8}\ & \underline{79.3}\ & \underline{87.1}\ & \underline{55.9}\ & 60\ & 98.3\ & \underline{83.7}\ \\
     GPT4-o  & 45.9\ & 77.0\ & 76.1\ & 84.4\ & 44.8\ & \underline{66.7}\ & 79.2\ & 73.7\ \\
     Seed-1.5-VL  & 37.4\ & 73.2\ & 75.4\ & 76.0 \ & 41.1\ & 46.7\ & \underline{\textbf{100}}\ & 36.8\ \\
     Seed-1.5-VL-Think  & 49.6\ & 80.4\ & 73.9\ & 82.7\ & 42.9\ & 46.7\ & 74.2\ & 75.1\ \\
     Qwen-2.5-VL-72B  & 36.7\ & 69.2\ &71.3\ & 65.1\ & 55.2\ & 26.7\ & 87.0\ & 40.0\ \\
     Qwen-2.5-VL-32B  & 43.3\ & 71.6\ & 60.5\ & 62.2\ & 48.0 \ & 26.7\ & 70.2\ & 37.0\ \\
     Qwen-2.5-VL-7B  & 31.1\ & 54.7\ & 37.5\ & 41.5\ & 20.5\ & 6.7\ & 47.5\ & 10.9\ \\
     GLM-4.1-9B-Think  & 34.1\ & 51.7\ & 22.8\ & 45.8\ & 14.0\ & 6.7\ & 86.0\ & 37.2\ \\
     RoboBrain-2.0-32B  & 29.6\ & 63.5\ & 58.2\ & 51.7\ & 41.2\ & 0.0\ & 43.6\ & 29.9\ \\
     RoboBrain-2.0-7B & 0.3\ & 31.4\ & 36.0\ & 33.1\ & 25.3\ & 0.0\ & 0.0\ & 22.2\ \\ \noalign{\vskip 1pt} \hline\noalign{\vskip 3pt} 
     \name{}-7B-SFT-wo-R & 55.2\ & 69.9\ & 82.5\ & 89.0\ & 91.5\ & 60.0\ & \textbf{100}\ & 90.5\ \\
     \name{}-7B-SFT & 57.8\ & 77.1\ & 85.8\ & 91.1\ & 84.2\ & 86.7\ & \textbf{100}\ & 88.4\ \\
     \name{}-7B-RL & 59.6\ & {85.4} & {93.2}\ & 90.3\ & 78.6\ & {86.7}\ & 95.9\ & 87.0\ \\
     \name{}-32B-SFT & {64.0}\ & {83.5}\ & {89.3}\ & 93.0\ & {89.7}\ & {80.0}\ & \textbf{100}\ & {95.1}\ \\
     \name{}-32B-RL & \textbf{64.4} & \textbf{86.8} & \textbf{96.6} & \textbf{96.0} & \textbf{92.5} & \textbf{93.3} & \textbf{100} & \textbf{96.2} \\
    \bottomrule
  \end{tabular}
  \caption{Offline evaluation results. \name{}-7B-SFT-wo-R refers to our SFT model without chain-of-thought reasoning, while \name{}-7B-RL denotes the full trained policy obtained by applying RL after SFT. For AGIBot, Internal OOD (Out-of-Distribution), and Internal ID (In Distribution)–MultiStage/Constrained/Interrupt/OpenEnded, we report plan accuracy; for Internal ID–Invalid/Replan, we report F1 score. The best result for each evaluation set is shown in \textbf{bold}, and the best among baselines is \underline{underlined}.} 
  \label{tab:offline_eval_results}
\end{table}

\paragraph{Evaluation metrics.}  For AGIBot, Internal OOD, Internal ID–MultiStage/Constrained/Interrupt/OpenEnded, we report action prediction accuracy by matching the predicted action against a candidate action list, with similarity judged by Seed-1.5-VL (see \Cref{appendix:prompts} for the prompt). For Internal ID–Invalid/Replan, which are binary classification tasks, we report the F1 score.

\paragraph{Baseline methods.} We compare against widely used commercial and open-source VLMs, including Gemini-2.5-Pro, GPT-4o, Seed-1.5-VL, Seed-1.5-VL-Think, Qwen2.5-VL-7B/32B/72B,  GLM-4.1V-9B-Thinking~\cite{hong2025glm}, and RoboBrain-2.0-7B/32B. All baselines are adapted to the multi-turn observation–think–action format using the prompts in \Cref{appendix:prompts}. For each model, we test both English and Chinese prompts and report the better result. Gemini-2.5-Pro and GPT-4o perform better with English prompts, whereas the other models achieve higher accuracy with Chinese prompts. All evaluations are conducted using greedy decoding.

\paragraph{Results.} The offline evaluation results for each model are presented in \Cref{tab:offline_eval_results}. Key observations include:
\begin{itemize}[leftmargin=0.5cm]
    \item \name{}-32B-RL ranks first on all evaluation sets, demonstrating strong task planning and human–robot interaction capabilities, and substantially outperforming all open-source and commercial VLMs on both ID and OOD benchmarks.
    \item Chain-of-thought reasoning is critical for both OOD generalization and complex instruction following. \name{}-7B-SFT without reasoning (\name{}-7B-SFT-wo-R) exhibits a drop of over 7 percentage points in accuracy on the Internal OOD benchmarks compared to its reasoning-enabled counterpart, and suffers a 26.7-point decline on the ID–OpenEnded tasks.
    \item RL is critical, boosting \name{}-32B’s performance on nearly all evaluation sets. On the challenging Internal OOD benchmarks, \name{}-7B-RL and \name{}-32B-RL improve accuracy by 8.3 and 3.3 points, respectively, compared to their SFT counterparts. As shown in the case study (\Cref{appx:case-study}), RL primarily enhances the SFT models by (i) reducing irrational reasoning steps, (ii) improving thought-action consistency, and (iii) minimizing formatting errors.
    \item Gemini-2.5-Pro is the strongest baseline, ranking first on most evaluation sets among baseline methods. Our evaluation suggests it is currently the leading foundation model for embodied AI applications.
\end{itemize}



\begin{figure}[t]
    \centering
    \includegraphics[width=0.8\linewidth]{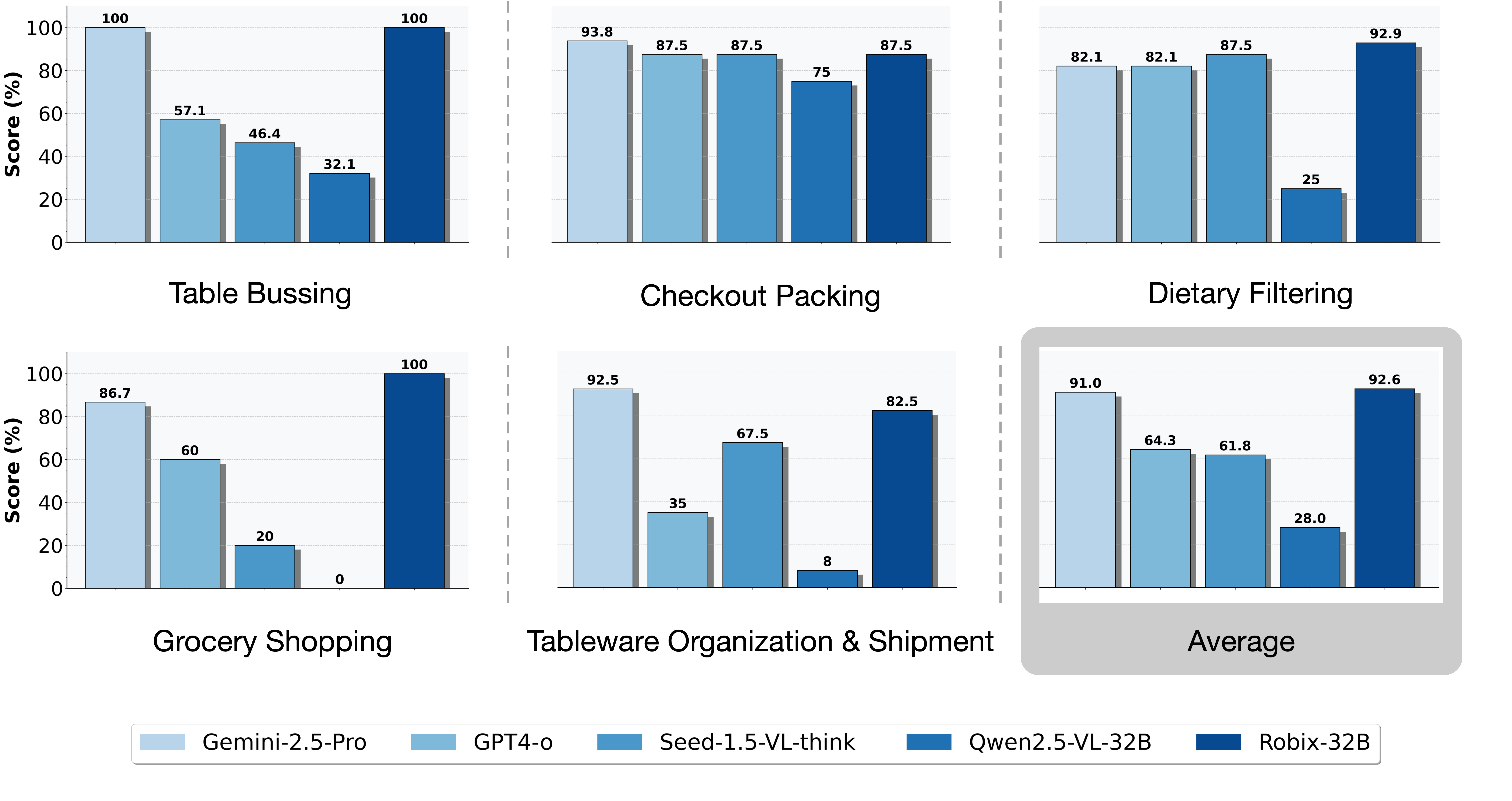}
    \caption{Online evaluation results with a human labeler operating a UMI device as the low-level controller.}
    \label{fig:online_exp}
\end{figure}

\subsection{Online Evaluation} 
\label{sec:online_exp}

While offline evaluation is cost-effective, it is limited to static environments and cannot assess a model’s ability to interact with the dynamic physical world. To address this, we deploy our model and baselines within a hierarchical robot system across diverse real-world settings—including kitchens, meeting rooms, and grocery stores—and conduct online evaluations to measure their effectiveness as high-level planning and interaction modules for daily tasks.
We design two sets of experiments:
\begin{itemize}[leftmargin=0.5cm]
    \item \textbf{Online evaluation of VLMs}. Assess the planning and interaction capabilities of VLMs in isolation, without the influence of low-level controllers.
    \item \textbf{Online evaluation of the VLM-VLA robot system}. Assess the end-to-end system performance by pairing the VLM with an automatic VLA model as the low-level controller.
\end{itemize}
In the first set of experiments, VLMs serve as the high-level planning and interaction module, while human labelers equipped with a Universal Manipulation Interface (UMI)~\cite{chi2024universal} device act as the low-level controller, enabling evaluation under a fully reliable control setting. In the second set, we use our in-house VLA model GR-3~\cite{cheang2025gr}, as the low-level controller and deploy the integrated VLM–VLA system on the ByteMini robot~\cite{cheang2025gr}. \name{} is deployed with customized inference optimization techniques~\cite{zheng2025tritondistributed, zheng2025tilelink} to reduce response latency.


\subsubsection{Online Evaluation of VLMs}
\label{sec:online_vlm}

\paragraph{Experimental settings.} We designed five tasks—table bussing, checkout packing, dietary filtering, grocery shopping, and tableware organization \& delivery—spanning diverse environments such as kitchens, meeting rooms, and grocery stores. To increase realism, some tasks deliberately incorporate user interruptions as well as fail-and-replan scenarios. For details on the initial states and brief descriptions of each task, please refer to \Cref{appendix:task_description}. We further annotated each task with subtasks to assess completion (e.g., for table bussing: ``tissue is in the rubbish bin'', ``plate is in the basket''). We compare \name{}-32B~\footnote{By default, \name{}-32B refers to \name{}-32B-RL.} with four baselines that performed well in the offline evaluation: Gemini-2.5-Pro, GPT-4o, Seed-1.5-VL-Think, and Qwen2.5-VL-32B. To reduce experimental variability, each task–model pair is repeated four times, and we report the average results. Following \citep{shi2025hi}, we use \textit{task progress}—the percentage of subtasks completed by the end of the task—as the evaluation metric. Trained human annotators assess task progress to ensure reliability and consistency.

\paragraph{Results.} The online evaluation results are shown in \Cref{fig:online_exp}. Both \name{}-32B and Gemini-2.5-Pro rank first in 3 of the 5 tasks, with \name{}-32B achieving a slightly higher average task progress (92.6\% vs. 91\%), demonstrating its superior performance in dynamic real-world environments. \name{}-32B also outperforms Qwen2.5-VL-32B by a large margin (92.6\% vs. 28\%), underscoring the effectiveness of our training pipeline. Gemini-2.5-Pro remains the strongest baseline, showing strong capabilities in following complex instructions. However, deploying large foundation models directly for planning and interaction introduces a major challenge—high response latency. In our experiments, Gemini-2.5-Pro sometimes required over 30 seconds to respond. While customized deployment may help reduce latency, we contend that current large-scale commercial VLMs remain too computationally heavy for real-time interaction, even on advanced hardware. 

\subsubsection{Online Evaluation of the VLM-VLA system}
\label{sec:real_exp}

\begin{figure}[t]
    \centering
    \includegraphics[width=0.8\linewidth]{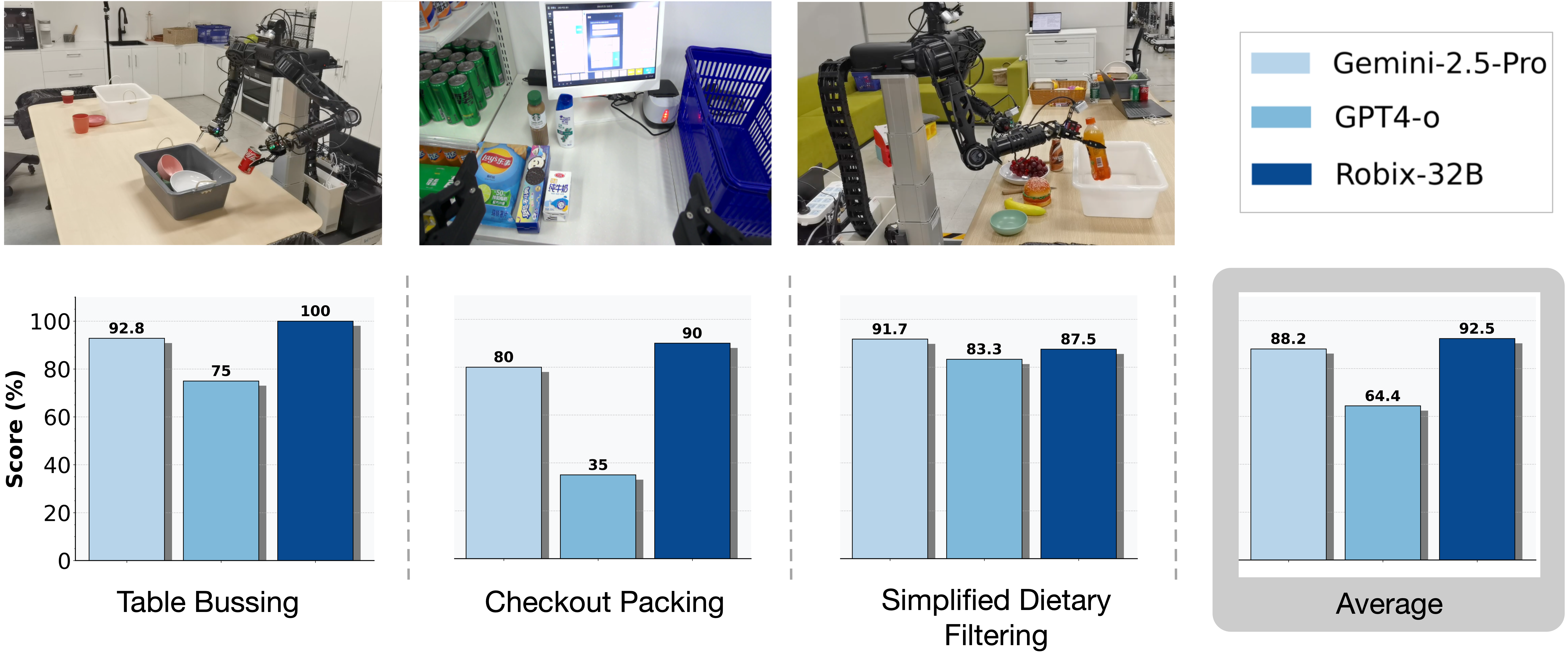}
    \caption{Online evaluation on the ByteMini robot with GR-3 model as the low-level controller.}
    \label{fig:online_exp_with_vla}
\end{figure}


\paragraph{Experimental Settings.} We select three tasks from the online evaluation—table cleaning, dietary filtering, and checkout packing—as the evaluation set, excluding the remaining two tasks that require actions beyond GR-3’s current capabilities. To better isolate the high-level cognitive layer’s performance, we also remove particularly challenging items to reduce frequent manipulation failures. Following the VLM online evaluation protocol, each task–model pair is evaluated four times, and we report average results using task progress as the metric. All experiments are conducted with the GR-3 model and the ByteMini robot~\cite{cheang2025gr}.

\paragraph{Results.} We compare \name{}-32B with the two strongest baselines from both offline and online evaluations: Gemini-2.5-Pro and GPT-4o. \Cref{fig:online_exp_with_vla} shows the results across the three real-world tasks. The findings mirror those in \Cref{sec:online_vlm}:  \name{}-32B achieves an average task progress of 92.5\%, exceeding Gemini-2.5-Pro and GPT-4o by 4.3 and 28.1 percentage points, respectively. We find that baseline methods—particularly GPT-4o—sometimes generate actions that are semantically correct but unrecognizable to the VLA. For instance, the VLA can execute “\textit{put the Oreo into the shopping basket}” but fails to interpret “\textit{put the biscuit box into the shopping basket}.” Such VLM–VLA misalignment mainly accounts for the online performance drop observed in Gemini-2.5-Pro and GPT-4o.

%% file: sections/5-relatedwork.tex
\section{Related Work}
\label{src:related_word}

\paragraph{Robotic Task Planning.}
Solving complex, long-horizon tasks in open environments demands robust high-level planning. Vision-Language Models (VLMs) have advanced robotic task planning by grounding high-level instructions in perceptual context~\cite{yang2024guiding,wang2024vlm,mei2024replanvlm}. Unlike Large Language Models (LLMs), which often generate ungrounded or physically infeasible plans due to a lack of environmental perception~\cite{huang2022inner,song2023llm}, VLMs integrate visual understanding with language reasoning to enable open-vocabulary instruction following and closed-loop planning. Systems such as COME-robot~\cite{zhi2024closed}, VILA~\cite{hu2023look}, and REPLAN~\cite{skreta2024replan} leverage GPT-4V to generate executable plans directly from raw visual observations and iteratively refine them based on environmental feedback, improving robustness through situated reasoning and failure recovery. Despite these advances, VLM-based approaches face persistent challenges: they struggle to maintain long-term consistency, exhibit limited embodied reasoning for grounding objects and spatial concepts in the physical world, and fail to fully integrate these signals for adaptive, task-centric planning. Addressing these issues is essential for scaling VLM-based planning to real-world, long-horizon embodied tasks. Moreover, most existing methods focus solely on task planning while overlooking the human–robot interaction capabilities during task execution that are crucial for a truly generalist robotic system.

\paragraph{Human-Robot Interaction.}
Existing work on human–robot interaction primarily focuses on enabling seamless, natural communication through real-time feedback and corrections. Early model-based systems grounded language in symbolic environment representations~\cite{swadzba2009computational,matuszek2013learning,namasivayam2023learning,patki2019inferring}, whereas recent learning-based methods adopt hierarchical architectures to directly interpret and act on user feedback~\cite{liu2023interactive,xiao2024robi,shi2024yell,belkhale2024rt,singh2024lgr2,mccallum2023feedback,driess2023palm,dai2024racer}. Examples include OLAF~\cite{liu2023interactive}, which uses GPT-4 to re-label actions and update visuomotor policies from corrections; YAY Robot~\cite{shi2024yell}, which integrates feedback into an iterative training loop but is limited by prompt diversity; RT-H~\cite{belkhale2024rt}, which supports language-based intervention but restricts corrections to fixed spatial moves; and RACER~\cite{dai2024racer}, which combines a VLM supervisor with physics simulation for recovery guidance. Hi Robot~\cite{shi2025hi} advances these approaches by grounding real-time corrections in the robot's own observations, enabling interpretation and execution of complex instructions beyond prior systems’ capabilities. However, achieving flexible interaction alongside adaptive task planning requires strong reasoning capacity—a challenge \name{} addresses by leveraging chain-of-thought reasoning to unify complex task planning and human–robot interaction within a single model. \name{} further introduces novel interaction capabilities, including proactive dialogue to clarify ambiguous instructions or infer user intent, and context-aware commonsense reasoning.

\paragraph{Embodied Reasoning.}
Embodied reasoning is the capacity of vision–language models (VLMs) to ground objects, spatial concepts, and physical interactions in the real world, and to integrate these signals into downstream robotic tasks~\cite{team2025gemini}. Unlike abstract symbolic reasoning, it is inherently action-oriented, requiring agents to interpret dynamic environments, plan context-aware behaviors, and adapt through feedback.
Recent advances span model design, data curation, and task-specific optimization. Embodied-Reasoner~\cite{zhang2025embodied} learns observation–thought–action trajectories enriched with spatial reasoning and self-reflection for visual search. Gemini Robotics-ER~\cite{team2025gemini} embeds embodied reasoning into its core VLM, achieving strong generalization across tasks such as 3D perception, pointing, state estimation, and affordance prediction.
Data-driven approaches include Cosmos-Reason1~\cite{azzolini2025cosmos}, which curates datasets emphasizing task-centric reasoning, and RoboBrain-2.0~\cite{team2025robobrain}, which synthesizes spatial–temporal reasoning datasets augmented with thought traces for causal chain learning.
Task-specific methods include EvolveNav~\cite{lin2025evolvenav}, which improves vision–language navigation via formalized CoT fine-tuning and self-reflective post-training; and ECoT~\cite{zawalski2024robotic}, which trains vision–language–action models for multi-step reasoning over plans, sub-tasks, motions, and grounded visual features before action generation. However, effectively leveraging embodied reasoning to develop generalist robotic systems capable of interactive, long-horizon task execution remains underexplored. \name{} addresses this gap by integrating robot reasoning, task planning, and natural language interaction to enable seamless interaction with both humans and physical environments, advancing toward general-purpose embodied intelligence.
 

%% file: sections/6-conclusion.tex
\section{Conclusion}
This paper presents \name{}, a unified vision-language model that integrates robot reasoning, adaptive task planning, and human-robot interaction. Serving as the high-level cognitive layer of a hierarchical robot system, \name{} enables robots to execute interactive, long-horizon tasks in open environments with high flexibility. It demonstrates flexible interaction capabilities, including proactive dialogue to clarify ambiguous instructions or infer user intent, real-time interruption handling, and context-aware commonsense reasoning. Experimental results show that \name{} delivers strong performance on real-world robotic tasks and exhibits robust generalization in out-of-distribution settings.

\paragraph{Limitations \& Future Work.}  Similar to other state-of-the-art multimodal models, \name{} has several limitations. In highly dynamic tasks with frequent scene transitions, it may produce hallucinations, flawed reasoning, or exhibit gaps in physical commonsense. Additionally, \name{} relies on short-term context windows to process interaction history, functioning as a form of short-term memory. Long-term interactive scenarios, however, require more advanced memory mechanisms—specifically, long-term memory with dynamic updates, efficient retrieval, and effective utilization, akin to context engineering in large language models. Addressing these challenges will be a primary focus of future work.

%% file: sections/7-contribution.tex
\section{Acknowledgements}

We thank Wanli Peng, Yongyu Yan, and Tingshuai Yan for their assistance with model deployment and inference optimization. We are also grateful to Baifeng Xie, Lihao Liu, and Yangang Zhang for their support in utilizing the internal simulation platform, and to Xiao Ma for his valuable suggestions on the writing of this paper. We further thank the GR-3 team for providing the teleoperation data, GR-3 model, and ByteMini robot used in our experiments. Finally, we thank the Seed-1.5-VL team for their support with data resources.

%% file: sections/appendix.tex
\section{Experiment details}

\subsection{Fundamental Evaluation Prompts}
\label{appx:base}
The prompt templates for all benchmarks are listed below. In each template, \texttt{\{question\}} is replaced with the sample’s actual question, \texttt{\{options\}} with its multiple-choice answer options, \texttt{<image>} with the computed ViT embeddings of the input image, and \texttt{<video>} with the ViT embeddings of the video frames (for video benchmarks).

\paragraph{VSIBench.} We use the official metric of VSIBench~\cite{yang2025thinking}.

\begin{tcolorbox}
[colback=white,colframe=black,left=1mm,right=1mm,top=1mm,bottom=1mm]
\ttfamily

<video>

\texttt{\{question\}}

Options: \texttt{\{options\}}

Answer with the option's letter from the given choices directly.
\end{tcolorbox}

\paragraph{BLINK.} 
Following RoboBrain 2.0~\cite{team2025robobrain}, we evaluate BLINK~\cite{fu2024blink} on two sub-tasks—spatial relation understanding and depth perception—and report the average accuracy across them.

\begin{tcolorbox}
[colback=white,colframe=black,left=1mm,right=1mm,top=1mm,bottom=1mm]
\ttfamily

<image>

\texttt{\{question\}}

\texttt{\{options\}}

 Answer with the option's letter from the given choices directly.
\end{tcolorbox}

\paragraph{CV-Bench.} We follow the official evaluation protocol of CV-Bench~\cite{tong2024cambrian} and report average accuracy over the three defined sub-tasks.

\begin{tcolorbox}
[colback=white,colframe=black,left=1mm,right=1mm,top=1mm,bottom=1mm]
\ttfamily

<image>

\texttt{\{question\}}

\texttt{\{options\}}

 Answer with the option's letter from the given choices directly.
\end{tcolorbox}

\paragraph{EmbSpatial.} We use the official metric of EmbSpatial-Bench~\cite{du2024embspatial} and report the average accuracy.

\begin{tcolorbox}
[colback=white,colframe=black,left=1mm,right=1mm,top=1mm,bottom=1mm]
\ttfamily

<image>

\texttt{\{question\}}

 Please answer with a single option letter:
 
 \texttt{\{options\}}

\end{tcolorbox}

\paragraph{SAT.}  We use the official metric of SAT~\cite{ray2024sat} and report the average accuracy across its five sub-tasks.
\begin{tcolorbox}
[colback=white,colframe=black,left=1mm,right=1mm,top=1mm,bottom=1mm]
\ttfamily

<image>

\texttt{\{question\}}
 
 \texttt{\{options\}}

 Please only answer with the option letter.

\end{tcolorbox}

\paragraph{VSR.} We use the official metric of VSR~\cite{liu2023vsr} and report the average accuracy.

\begin{tcolorbox}
[colback=white,colframe=black,left=1mm,right=1mm,top=1mm,bottom=1mm]
\ttfamily

<image>

\texttt{\{question\}}

Answer with a single word yes or no.

\end{tcolorbox}

\paragraph{SpatialBench.} 
We use the official metric of SpatialBench~\cite{du2024embspatial}.

\begin{tcolorbox}
[colback=white,colframe=black,left=1mm,right=1mm,top=1mm,bottom=1mm]
\ttfamily

<image>

\{question\}

Please answer with a single word: yes or no (for existence task).

\end{tcolorbox}

\paragraph{DA-2k.} Since some VLMs tend to repeatedly output the same option on DA-2k~\cite{yang2024depth}, we shuffle the answer choices and report the average accuracy.

\begin{tcolorbox}
[colback=white,colframe=black,left=1mm,right=1mm,top=1mm,bottom=1mm]
\ttfamily

<image>

\texttt{\{question\}}
 
Only provide the answer: point1 or point2.

\end{tcolorbox}

\paragraph{LVIS.} We report results on LVIS~\cite{gupta2019lvis} using the F1-score~\footnote{Since the baseline models (\eg{}, Qwen-2.5-VL, Cosmos-Reason1, RoboBrain-2.0, etc.) cannot follow a unified output format on the visual grounding benchmarks—including LVIS, RefCOCO, Pixmo-Point, and Where2Place—we revise their prompts to adapt to their output formats.}.
\begin{tcolorbox}
[colback=white,colframe=black,left=1mm,right=1mm,top=1mm,bottom=1mm]
\ttfamily
<image>

Ground all \texttt{\{object\}} in this image. Please provide all bounding box coordinates in the format: <|box\_start|>(x1, y1),(x2, y2)<|box\_end|>.

\vspace{1em}

Each bounding box should contain: 

- (x1, y1): the coordinates of the upper-left corner

- (x2, y2): the coordinates of the lower-right corner 

\vspace{1em}

All coordinates must be **normalized to the range [0, 1000]**, where:

- x refers to the horizontal axis (image width)

- y refers to the vertical axis (image height) 

\vspace{1em}

If **no object is detected**, simply output:  **No instance found.**
\end{tcolorbox}

\paragraph{RefCOCO \& RefCOCO+ \& RefCOCOg \& RefCOCOu.} We use the official metric of RefCOCO~\cite{yu2016modeling}.
\begin{tcolorbox}
[colback=white,colframe=black,left=1mm,right=1mm,top=1mm,bottom=1mm]
\ttfamily
<image>

What are the coordinates of the \{object\} in this image?

Please provide the bounding box coordinates of the \{object\} in the format: <|box\_start|>(x1, y1),(x2, y2)<|box\_end|>, where (x1, y1) for upper-left, and (x2, y2) for lower-right. All coordinates should be normalized to a [0, 1000] scale, where x corresponds to the horizontal axis (image width), y corresponds to the vertical axis (image height).
\end{tcolorbox}

\paragraph{Pixmo-Point \& Where2Place.} For these two benchmarks~\cite{deitke2024molmo,yuan2024robopoint}, we compute the proportion of predicted points that fall within the referring objects.
\begin{tcolorbox}
[colback=white,colframe=black,left=1mm,right=1mm,top=1mm,bottom=1mm]
\ttfamily
<image>

Locate several points for the \{object\}.

You can mark them using <point>x y</point>.

Please provide point coordinates in the format: <point>(x1, y1)</point> ... 

The coordinates of the point (x, y) must be **normalized to the range [0, 1000]**, where:

- x refers to the horizontal axis (image width) 
- y refers to the vertical axis (image height)

\end{tcolorbox}

\paragraph{VisualWebBench.} We use the official metric of VisualWebBench~\cite{liu2024visualwebbench}.

\begin{tcolorbox}
[colback=white,colframe=black,left=1mm,right=1mm,top=1mm,bottom=1mm]
\ttfamily

<image>

\texttt{\{question\}}

\end{tcolorbox}

\paragraph{ERQA.} We report accuracy on ERQA~\cite{team2025gemini}.

\begin{tcolorbox}
[colback=white,colframe=black,left=1mm,right=1mm,top=1mm,bottom=1mm]
\ttfamily

<image>...<image>

\texttt{\{question\}}

\{options\}

Please answer directly with only the letter of the correct option and nothing else.

\end{tcolorbox}

\paragraph{RoboVQA \& EgoTaskQA \& OpenEQA-hm3d \& OpenEQA-scannet.} We compute accuracy by using GPT-4o to compare model responses with the ground-truth labels, and the evaluation prompt is shown below. 

\begin{tcolorbox}
[breakable,colback=white,colframe=black,left=1mm,right=1mm,top=1mm,bottom=1mm,fontupper=\small]
\ttfamily

<video>

\texttt{\{question\}}

\end{tcolorbox}

\begin{tcolorbox}
[breakable,colback=white,colframe=black,left=1mm,right=1mm,top=1mm,bottom=1mm,fontupper=\small]
\ttfamily

\textbf{Prompt for GPT-4o (RoboVQA \& EgoTaskQA).}

You are an AI assistant tasked with evaluating whether a response matches the correct answer to a given question.
\vspace{1em}

\textbf{Evaluation Rules}

(1) Output 1 if the response matches the answer exactly or with synonymous/equivalent wording.

- Synonyms, paraphrases, or different surface forms of the same meaning count as matches.

- Minor wording differences (e.g., ``put tomato into fridge'' vs. ``the person is putting a tomato in the fridge'') count as matches.

\vspace{1em}

(2) Output 0 if the response is incorrect, contradictory, or refers to a different entity, object, or attribute.

- If the answer and response describe different objects, actions, or states, mark as 0.

- If the response introduces additional details that change the meaning of the answer, mark as 0.
\vspace{1em}

\textbf{Special Cases}

- Similar meaning: Output 1 if the response conveys essentially the same meaning as the answer and does not omit or add critical information (e.g., answer:``put meat on the table'', response:``The person moved meat from the fridge to the counter.'').

- Partial matches: If the response overlaps but misses or alters essential details (e.g., answer:``put meat and tomato on the table'' vs. response:``put meat on the table''), output 0.

- Granularity differences: If the response is more specific but still semantically equivalent (e.g., answer:``woman'', response:``Jessica''), output 1.

- Yes/No questions: Only output 1 if the polarity matches (yes <-> yes, no <-> no). Any mismatch outputs 0, regardless of explanation.

- Ambiguity: If the response cannot be reasonably interpreted as equivalent to the answer, output 0.

\vspace{1em}

\textbf{Examples}

Example 1

Question: Did the attribute of plant changed because of the action getting something from something?

Answer: yes

Response: Yes, the attribute of plant got watered from no to yes after the action getting something from something.

Your output: 1

\vspace{1em}

Example 2
Question: what status of fork changed while the person do the first action did before he/she put something to something?

Answer: cleanliness

Response: fork was in drawer before the person put fork to sink.

Your output: 0

\vspace{1em}

Example 3

Question: What is the person doing before he/she close something?

Answer: Put tomato to fridge

Response: The person is putting a tomato in the fridge.

Your output: 1

\vspace{1em}

Example 4

Question: What is the first action the person did in the video?

Answer: Work on sofa

Response: The person pulled out a chair.

Your output: 0

\vspace{1em}

Example 5

Question: How did the person changed the spatial relationships of meat?

Answer: Put meat to table

Response: The person moved meat from the fridge to the counter.

Your output: 1

\vspace{1em}

Example 6

Question: what status of fridge changed while the person do the first action did after he/she point to something?

Answer: openess

Response: The fridge was closed before the person point to something, and after that the fridge changed to open.

Your output: 1

\vspace{1em}

Example 7

Question: which object changed its status when the person do the last action in the video?

Answer: fork

Response: spoon

Your output: 0

\vspace{1em}

Example 8

Question: What is the action that just happened?

Answer: Place can in the tray

Response: The person puts the can on the table.

Your output: 0

\vspace{1em}

Example 9

Question: current goal is: Please place the fruits in the bowl then place the kitchen supplies into the holder. last 20 steps: 1. put white packet in the bowl 2. put white packet in the bowl 3. put yellow packet in the bowl 4. put blue packet in the bowl 5. put blue packet in the bowl 6. put blue packet in the bowl 7. put yellow packet in the bowl. What's the immediate next step?

Answer: Put duster in the black stand

Response: put brush in the holder

Your output: 0

\vspace{1em}

Your Turn:

Question: \{question\}

Answer: \{answer\}

Response: \{prediction\}

Your output:
\end{tcolorbox}

\begin{tcolorbox}
[breakable,colback=white,colframe=black,left=1mm,right=1mm,top=1mm,bottom=1mm,fontupper=\small]
\ttfamily

\textbf{Prompt for GPT-4o (OpenEQA-hm3d \& OpenEQA-scannet).}
You are an AI assistant tasked with evaluating whether a response matches the correct answer to a given question, considering both the primary answer and any extra correct answers.

\vspace{1em}

\textbf{\#\#Evaluation Rules}

(1) Output 1 if the response matches the answer or any of the extra answers exactly or with synonymous/equivalent wording.

- Synonyms, paraphrases, or different surface forms of the same meaning count as matches.

- Minor wording differences (e.g., ``Wood panel'' vs. ``Wood) count as matches.

(2) Output 0 if the response is incorrect, contradictory, or refers to a different entity, object, or attribute than the answer and all extra answers.

- If the answer and response describe different objects, actions, or states, mark as 0.

- If the response introduces additional details that change the meaning of the answer, mark as 0.

\vspace{1em}

\textbf{\#\#Special Cases}

- Similar meaning: Output 1 if the response conveys essentially the same meaning as the answer and does not omit or add critical information (e.g., answer: ``A ceiling fan'', response: ``fan'').

- Partial matches: If the response overlaps but misses or alters essential details (e.g., answer: ``put meat and tomato on the table'' vs. response: ``put meat on the table''), output 0.

- Granularity differences: If the response is more specific but still semantically equivalent (e.g., answer: ``woman'', response: ``Jessica''), output 1.

- Yes/No questions: Only output 1 if the polarity matches (yes <-> yes, no <-> no). Any mismatch outputs 0, regardless of explanation.

- Ambiguity: If the response cannot be reasonably interpreted as equivalent to the answer, output 0.

\vspace{1em}

\textbf{\#\#Examples:}

Example 1:

Question: Is it overcast?

Answer: no

Extra Answers: ["doesn't look like it", "no", "it's sunny"]

Response: yes

Your output: 0

\vspace{1em}

Example 2:

Question: Who is standing at the table?

Answer: woman

Extra Answers: ["a woman", "a lady", "woman"]

Response: Jessica

Your output: 1

\vspace{1em}

Example 3:

Question: Are there drapes to the right of the bed?

Answer: yes

Extra Answers: ["yes, there are drapes", "yeah", "the drapes are to the right of the king bed"]

Response: yes

Your output: 1

\vspace{1em}

Example 4:

Question: What material is the ceiling in the living room?

Answer: Wood panel

Extra Answers: null

Response: wood

Your output: 1

\vspace{1em}

Example 5:

Question: What is in between the two picture frames on the blue wall in the living room?

Answer: The TV

Extra Answers: null

Response: air conditioner

Your output: 0

\vspace{1em}

Example 6:

Question: Is the house doorway open or closed?

Answer: Open

Extra Answers: null

Response: The house doorway is open.

Your output: 1

\vspace{1em}

Example 7:

Question: Is my backyard safe to let me dog out in?

Answer: Yes, its fenced.

Extra Answers: null

Response: yes

Your output: 1

\vspace{1em}

Example 8:

Question: What is hanging from the ceiling in the bedroom?

Answer: A ceiling fan

Extra Answers: null

Response: fan

Your output: 1

\vspace{1em}

Example 9:

Question: Where is the full body mirror?

Answer: In the bedroom by the door

Extra Answers: ["next to the bedroom door", "just inside the bedroom", "in the bedroom", "in the bedroom right next to the door"]

Response: The full body mirror is in the bedroom.

Your output: 1

\vspace{1em}

Example 10:

Question: What is leaning in the corner by the coat rack?

Answer: An umbrella

Extra Answers: null

Response: chair

Your output: 0

\vspace{1em}

Your Turn:

Question: \{question\}

Answer: \{answer\}

Extra Answers: \{extra\_answers\}

Response: \{prediction\}

Your output: 

\end{tcolorbox}










\paragraph{AgiBot-ER.} We curate 97, 120, 381 test samples for three sub-tasks, $i.e.$, Task Status Verification, Action Affordance, and Next Action Prediction, respectively.  We report the average accuracy of the three tasks.


\begin{tcolorbox}
[colback=white,colframe=black,left=1mm,right=1mm,top=1mm,bottom=1mm]
\ttfamily
\textbf{Task Status Verification}

<image><image><image><image><image><image>

\texttt{\{question\}}

\texttt{Your answer can only be "yes" or "no"}
\vspace{1em}

\textbf{Action Affordance}

<image><image><image>

\texttt{\{question\}}

\texttt{Your answer can only be "yes" or "no"}.
\vspace{1em}

\textbf{Next Task Prediction}

<image><image><image><image><image><image>

\texttt{\{question\}}

\{options\}

\texttt{Answer with the option’s letter from the given choices directly.}



\end{tcolorbox}

\paragraph{MME.} We report the \textit{accuracy+} metric for MME~\cite{fu2024mme}.

\begin{tcolorbox}
[colback=white,colframe=black,left=1mm,right=1mm,top=1mm,bottom=1mm]
\ttfamily

<image>

\texttt{\{question\}}

Please answer yes or no.

\end{tcolorbox}

\paragraph{MMBench.} We demonstrate the official metric provide by MMBench~\cite{liu2024mmbench}.

\begin{tcolorbox}
[colback=white,colframe=black,left=1mm,right=1mm,top=1mm,bottom=1mm]
\ttfamily

<image>

\texttt{\{question\}}

\{options\}

Answer with the option's letter from the given choices directly.

\end{tcolorbox}

\paragraph{RealWorldQA.} We use the official metric of RealWorldQA~\cite{realworldqa}.

\begin{tcolorbox}
[colback=white,colframe=black,left=1mm,right=1mm,top=1mm,bottom=1mm]
\ttfamily

<image>

\texttt{\{question\}}

\{options\}

Please answer directly with only the letter of the correct option and nothing else.

\end{tcolorbox}

\paragraph{SimpleVQA.} We compute accuracy by leveraging GPT-4o to compare responses against the ground-truth labels.

\begin{tcolorbox}
[colback=white,colframe=black,left=1mm,right=1mm,top=1mm,bottom=1mm]
\ttfamily

<image>

\texttt{\{question\}}

\end{tcolorbox}

\paragraph{EgoSchema \& VideoMME \& NextQA.} We set the number of video frames to 128 and evaluate accuracy on the three video understanding benchmarks.

\begin{tcolorbox}
[colback=white,colframe=black,left=1mm,right=1mm,top=1mm,bottom=1mm]
\ttfamily

<video>

\texttt{\{question\}}

\{options\}

Answer with the option's letter from the given choices directly.

\end{tcolorbox}














\paragraph{MathVista.} We use the official metric of MathVista~\cite{lu2023mathvista}.

\begin{tcolorbox}
[colback=white,colframe=black,left=1mm,right=1mm,top=1mm,bottom=1mm]
\ttfamily

<image>

\texttt{\{question\}}

\end{tcolorbox}

\paragraph{MathVision.} We use the official metric of MathVision~\cite{wang2024measuring}.

\begin{tcolorbox}
[colback=white,colframe=black,left=1mm,right=1mm,top=1mm,bottom=1mm]
\ttfamily

<image>

\texttt{\{question\}}

\{options\}

Please only output your final answer or a single letter (if it is a multiple choice question).

\end{tcolorbox}

\paragraph{MMMU.} We use the official metric of MMMU~\cite{yue2024mmmu}.

\begin{tcolorbox}
[colback=white,colframe=black,left=1mm,right=1mm,top=1mm,bottom=1mm]
\ttfamily

<image>

\texttt{\{question\}}

\{options\}

Answer with the option's letter from the given choices directly.

\end{tcolorbox}

\subsection{The AGIBot offline evaluation set}
\label{appx:agibot}
We select 16 high-frequency daily tasks from the AGIBot dataset for offline evaluation: ``Heat food in the microwave oven,'' ``Make a sandwich,'' ``Wash dishes with dishwasher,'' ``Wash the dishes and silverware,'' ``Discard the trash on the coffee table,'' ``Pick up the item to wipe away the stain,'' ``Place items from the meeting room table into the storage box,'' ``Arrange sofa,'' ``Wash clothes in the washing machine,'' ``Remove clothes from the washing machine,'' ``Arrange flowers,'' ``Boil water in the kettle,'' ``Packing in the supermarket,'' ``Scan and package the goods,'' ``Restock the hanging basket area,'' and ``Restock supermarket snacks.'' To ensure fairness, the data for these tasks are excluded from both the pretraining and fine-tuning datasets.

\subsection{Offline evaluation examples}
\label{appendix:offline_example}

We present a snapshot of our offline evaluation sets.

\begin{tcolorbox}
[breakable,boxrule=0pt,colback=white,left=1mm,right=1mm,top=1mm,bottom=1mm,fontupper=\small]
\ttfamily
[

    \hspace{2em}...

    \hspace{2em} \{
    
    \hspace{4em}    "role": "user",
        
    \hspace{4em}    "content": "<|system\_info\_start|>Your current views are<image><image><image><|system\_info\_end|><|human\_start|>I want some fruit, put all fruits into the basket.<|human\_end|>"
        
    \hspace{2em} \},
    
    \hspace{2em} \{
        
        \hspace{4em} "role": "assistant",
        
        \hspace{4em} "content": "<|think\_start|>...<|think\_end|><|plan\_start|>put the banana into the plastic box<|plan\_end|><|response\_start|>...<|response\_end|>",
        
        \hspace{4em}"other\_actions": [
            "put the grapes into the plastic box"
        ]
        
    \hspace{2em} \},

    \hspace{2em} \{
    
    \hspace{4em}    "role": "user",
        
    \hspace{4em}    "content": "<|system\_info\_start|>Your current views are<image><image><image><|system\_info\_end|><|human\_start|>Wait, I don't want banana.<|human\_end|>"
        
    \hspace{2em} \},

    \hspace{2em} \{
        
        \hspace{4em} "role": "assistant",
        
        \hspace{4em} "content": "<|think\_start|>...<|think\_end|><|plan\_start|>put the banana on the table<|plan\_end|><|response\_start|>...<|response\_end|>",
        
        \hspace{4em}"other\_actions": [
        ]
        
    \hspace{2em} \},
    
    \hspace{2em}...
    
]
\end{tcolorbox}

The above example illustrates two consecutive rounds  within a trajectory. We query VLMs in a manner analogous to multi-turn dialogue interactions.
For the first round, planning is considered correct if the VLM output semantically matches either: \texttt{<|plan\_start|>put the banana into the plastic box<|plan\_end|>} or \texttt{<|plan\_start|>put the grapes into the plastic box<|plan\_end|>}. 
For the second round, we evaluate models under a teacher-forcing setting, incorporating the ground-truth from the first round into the dialogue history. Since the other\_actions list is empty for this step, planning is deemed correct only if the VLM output semantically matches:
\texttt{<|plan\_start|>put the banana on the table<|plan\_end|>}.

\subsection{Online evaluation tasks}
\label{appendix:task_description}

\begin{figure}[t]
    \centering 
    
    \begin{subfigure}{0.19\textwidth}
        \centering
        \includegraphics[width=\textwidth]{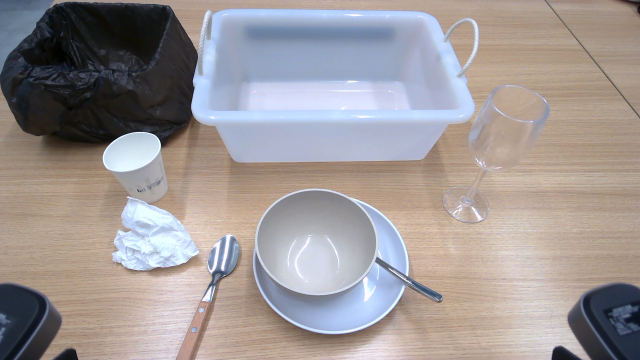}
        \caption{}
        \label{subfig:table_bussing}
    \end{subfigure}
    \begin{subfigure}{0.19\textwidth}
        \centering
        \includegraphics[width=\textwidth]{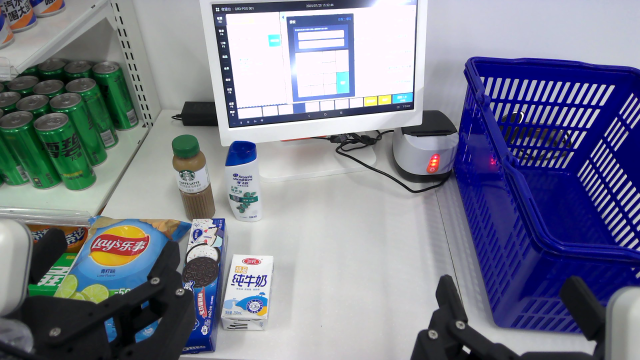}
        \caption{}
        \label{subfig:1}
    \end{subfigure}
    \begin{subfigure}{0.19\textwidth}
        \centering
        \includegraphics[width=\textwidth]{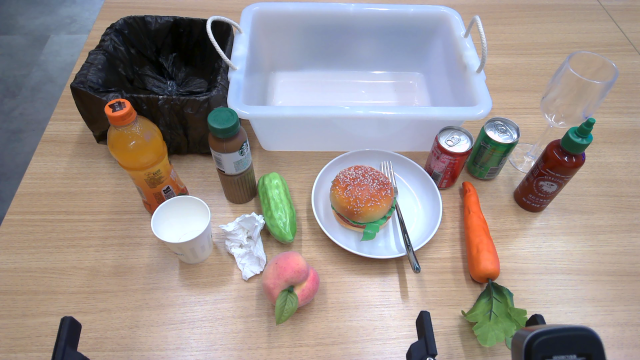}
        \caption{}
        \label{subfig:1}
    \end{subfigure}
    \begin{subfigure}{0.19\textwidth}
        \centering
        \includegraphics[width=\textwidth]{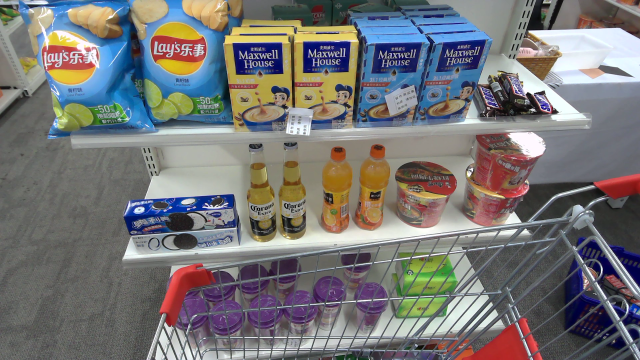}
        \caption{}
        \label{subfig:1}
    \end{subfigure}
    \begin{subfigure}{0.19\textwidth}
        \centering
        \includegraphics[width=\textwidth]{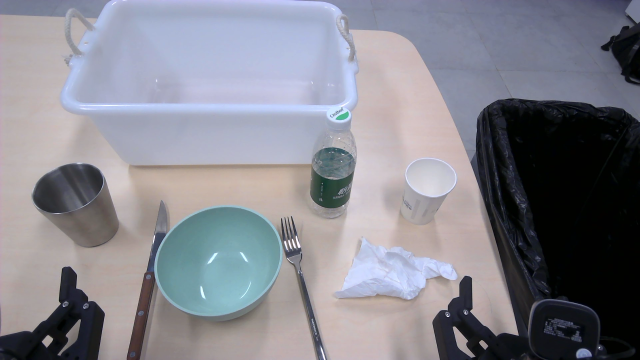}
        \caption{}
        \label{subfig:shipment}
    \end{subfigure}
    \caption{Example initial states of the five tasks used in our online evaluation. (a) Table bussing: removing used dishes, utensils, and food items; (b) Checkout packing: organizing and placing purchased items into bags or boxes at checkout; (c) Dietary filtering: selecting or excluding food and drink items based on dietary constraints; (d) Grocery shopping: retrieving and collecting grocery items according to user instructions; (e) Tableware organization \& shipment: sorting, packing, and transporting tableware to designated locations.}
    \label{fig:online_eval_tasks} 
\end{figure}

Example initial states of the five tasks used in our online evaluation are shown in \Cref{fig:online_eval_tasks}. Below, we provide an overview of each task.
\begin{itemize}[leftmargin=0.5cm]
    \item \textbf{Table bussing}: removing used dishes, utensils, and food items. In our experiment, the user requests the robot to clear the table, but interrupts with the command ``\textit{I still need it}'' while the robot is placing a glass into the plastic box. We also introduce a failure event when the robot attempts to place a spoon into the box.
    \item \textbf{Checkout packing}: organizing and placing purchased items into bags or boxes at checkout. In our experiment, the user asks the robot to move all items on the checkout counter into the shopping basket. An interruption occurs with the command ``\textit{I am allergic to caffeine}'' while the robot is placing a coffee into the basket.
    \item \textbf{Dietary filtering}: selecting or excluding food and drink items based on dietary constraints (e.g., no caffeine, vegetarian). In our experiment, the user issues a series of instructions: (1) ``\textit{Put the food with the highest energy into the plastic box, then add a drink. Note that I am allergic to caffeine.}''; (2) ``\textit{Discard all drinks containing caffeine into the rubbish bin}''; (3) ``\textit{Place all fruits and vegetables into the plastic box}''; and (4) ``\textit{Clean up all remaining items on the table}''.
    \item \textbf{Grocery shopping}: retrieving and collecting grocery items according to user instructions. In our experiment, the user first asks the robot to recommend a snack suitable for a road trip. Following the recommendation, the user requests a less sweet snack for the cart, then asks for a non-alcoholic drink, and finally makes a conditional request: ``Are there any nuts on the shelf? If so, put some into the shopping cart.''
    \item \textbf{Tableware organization \& shipment} : sorting, packing, and transporting tableware to designated locations. In our experiment, the user instructs the robot to discard all trash and transport the tableware to the sink. Specifically, the robot must first dispose of garbage in the rubbish bin, then place all tableware into a plastic box, and finally carry the box to the sink. During this process, the user interrupts with the command ``\textit{Keep it on the table}'' while the robot is placing a stainless steel cup into the box.
\end{itemize}

\subsection{Offline \& Online Evaluation Prompts}

\label{appendix:prompts}

We present the prompts used in our experiments, including the judge prompt for offline evaluation, the baseline model prompts, and the reward model prompt for reinforcement learning.

\textbf{Judge Prompt for Offline Evaluation}
\begin{tcolorbox}
[breakable,colback=white,colframe=black,left=1mm,right=1mm,top=1mm,bottom=1mm,fontupper=\small]
\ttfamily

\# Role

You are a judge to decide the similarity of two sentences.

\vspace{1em}

\# Inputs

You will be provided with two sentences, each sentence represent an action from a robot.

\vspace{1em}

\# Output Format

Your output should be a single number that is either 0 or 1, where 0 means the two sentences are semantically different, and 1 means the two sentences are semantically the same. Output in the following format: "[SOME ANALYSIS]. The final answer is: [NUMBER]"

\vspace{1em}

\# Examples

User:

1. put the fork into the basket

2. put the fork into the basket

Assistant:

Two sentences are exactly the same. The final answer is: 1.0

\vspace{1em}

User:

1. put the iron fork into the woven basket

2. put the fork into the basket

Assistant:

Two sentences are the same semantically. The final answer is: 1.0

\vspace{1em}

User:

1. Grab the bottled french fries on the table with the right arm.

2. Grab the bottled french fries.

Assistant:

Two sentences are the same semantically. The final answer is: 1.0

\vspace{1em}

User:

1. Move forward slightly

2. Move forward

Assistant:

Two sentences are the same semantically. The final answer is: 1.0

\vspace{1em}

User:

1. put the coffee into the plastic box

2. put the bottle of coffee into the plastic box

Assistant:

Two sentences are the same semantically. The final answer is: 1.0

\vspace{1em}

User:

1. put the glass into the plastic box

2. put the glass cup into the plastic box

Assistant:

Two sentences are the same semantically. The final answer is: 1.0

\vspace{1em}

User:

1. put the fork into the basket

2. put the knife into the basket

Assistant:

The objects (fork v.s. knife) are different semantically. The final answer is: 0.0

\vspace{1em}

User:

1. put the fork into the basket

2. pick up the fork

Assistant:

The actions are different semantically. The final answer is: 0.0

\vspace{1em}

User:

1. put the fork into the basket

2. navigate to the table

Assistant:

Neither actions nor objects matches. The final answer is: 0.0

\vspace{1em}

User:

1. put the fork into the basket

2. pick up the stainless steel cup

Assistant:

Neither actions nor objects matches. The final answer is: 0.0

\end{tcolorbox}

\textbf{English Prompt for Baseline Models.}

We use the following prompt to configure VLMs as the interaction and planning modules of the robot system in both offline and online experiments.
\begin{tcolorbox}
[breakable,colback=white,colframe=black,left=1mm,right=1mm,top=1mm,bottom=1mm,fontupper=\small]
\ttfamily
\# Role

You are a robot with two grippers developed by Bytedance, and your name is Roobio. You are deployed in a home environment, and your job is to have natural interactions with users and complete some tasks required by the user.

You have basic movement, perception, and manipulation capabilities, and can navigate to the following areas: dining table, sink, refrigerator, dishwasher, microwave, bread maker, cupboard, shoe cabinet, sofa, and washing machine.

\vspace{1em}

\# Input

In each round, you will be provided with the following information.

1. Three images: captured by the cameras from your head/left gripper/right gripper respectively. These images reflect your current visual perception at the current moment.

2. User instructions (optional): user may give you some instructions, user instructions are in chinese.

\# Output

To distinguish reasoning, planning, response and query. You must output with the following fields.

1. `<|think\_start|> ... <|think\_end|>': this field is for your reasoning process, it can include

\hspace{2em}    - The key items in the current scene.

\hspace{2em}    - Whether the previous action is complete.
    
\hspace{2em}    - Reasoning about your next action.

2. `<|plan\_start|> ... <|plan\_end|>' (optional): this field is for task planning, You should plan your next action based on your reasoning process. The action should be helpful to complete user's task.

3. `<|response\_start|> ... <|response\_end|>' (optional): this field is your response to user, you response must be in *chinese*. You should respond to the user in the following situations.
    
\hspace{2em}    - When the user is chatting with you;
    
\hspace{2em}    - When the user gives a new instruction on completing a task, you should always respond to the user whenever you receive a new instruction. The response should be concise, polite, and relevant to the task.
    
\hspace{2em}    - When you need to ask the user for more information. For example, if the user ask you to give him/her a drink, and you observe both orange juice and apple juice, then you should ask the user which one he/she prefers. You should avoid asking the user for more information too frequently.

\vspace{1em}

\# Guidelines

1. The text inside `<|think\_start|> ... <|think\_end|>' and `<|plan\_start|> ... <|plan\_end|>' should be in english while the text in `<|response\_start|> ... <|response\_end|>' must be in chinese.

2. When the user interrupt your action, you need to stop the current action and re-plan.

3. Your task planning inside `<|plan\_start|> ... <|plan\_end|>' should be in the following format:

\hspace{2em}    - put the ... into the ...

\hspace{2em}    - pick up the ...

\hspace{2em}    - navigate to the ...

\hspace{2em}    - open the ...

\hspace{2em}    - close the ...

4. Only include one action at a time, planning such as `<|plan\_start|> put the A into the B, put the C into the B<|plan\_end|>' is invalid.

\vspace{1em}

\# Examples

User:

<|system\_info\_start|>Your current views are...<|system\_info\_end|><|human\_start|>Robot, clean up the table for me.<|human\_end|>

Assistant:

<|think\_start|>The user asks me to clean up the table, I can observe.... <|think\_end|><|plan\_start|>put the ... into the ...<|plan\_end|><|response\_start|>...<|response\_end|>

User:

<|system\_info\_start|>The views you see after finishing previous action are...<|system\_info\_end|>

Assistant:

<|think\_start|>I should continue cleaning up the table, there are ... left on the table, ...<|think\_end|><|plan\_start|>put the ... into the ...<|plan\_end|>

...
\end{tcolorbox}

\textbf{Chinese Prompt for Baseline Models}

We also developed a corresponding Chinese prompt for models that perform better with Chinese.
\begin{tcolorbox}
[breakable,colback=white,colframe=black,left=1mm,right=1mm,top=1mm,bottom=1mm,fontupper=\small]
\ttfamily
\begin{CJK*}{UTF8}{gbsn}
    \texttt{\# 角色}
\end{CJK*}

\begin{CJK*}{UTF8}{gbsn}
    \texttt{你是字节跳动研发的双臂机器人。你被部署在一个开放式厨房环境中，负责与人类用户进行自然互动并完成日常操作任务。你具备基础的移动、感知与操控能力，可前往以下区域执行操作：餐桌旁、水池旁、冰箱旁、洗碗机旁、微波炉旁、烤面包机旁、水池旁的台面、鞋柜旁、沙发旁和洗衣机旁。}
\end{CJK*}

\vspace{1em}

\begin{CJK*}{UTF8}{gbsn}
    \texttt{\# 输入信息}
\end{CJK*}

\begin{CJK*}{UTF8}{gbsn}
    \texttt{在每轮交互中，你将接收到以下信息：}
\end{CJK*}

\begin{CJK*}{UTF8}{gbsn}
    \texttt{（1）三张图像：分别来自你头部摄像头、左夹爪摄像头、右夹爪摄像头，反映你在当前时刻的视觉感知。}
\end{CJK*}

\begin{CJK*}{UTF8}{gbsn}
    \texttt{（2）用户指令（可选）：人类用户可能通过语音或文本向你发出任务指令。}
\end{CJK*}

\vspace{1em}

\begin{CJK*}{UTF8}{gbsn}
    \texttt{\# 输出格式}
\end{CJK*}

\begin{CJK*}{UTF8}{gbsn}
    \texttt{你必须输出以下结构化内容，用于指导任务执行与用户交互。输出字段包括：}
\end{CJK*}

\begin{CJK*}{UTF8}{gbsn}
    \texttt{（1）<|think\_start|> ... <|think\_end|>：你的内部思考与任务状态判断，包括：}
\end{CJK*}

\begin{CJK*}{UTF8}{gbsn}
    \texttt{\hspace{2em}- 当前场景中包含的关键物品；}
\end{CJK*}

\begin{CJK*}{UTF8}{gbsn}
    \texttt{\hspace{2em}- 是否完成上一步操作；}
\end{CJK*}

\begin{CJK*}{UTF8}{gbsn}
    \texttt{\hspace{2em}- 当前场景变化的观察；}
\end{CJK*}

\begin{CJK*}{UTF8}{gbsn}
    \texttt{\hspace{2em}- 下一步应执行的关键动作与理由；}
\end{CJK*}

\begin{CJK*}{UTF8}{gbsn}
    \texttt{\hspace{2em}- 如遇异常（如抓取失败），应在此说明并调整策略。}
\end{CJK*}

\begin{CJK*}{UTF8}{gbsn}
    \texttt{（2）<|plan\_start|> ... <|plan\_end|>（可选）：你对下一步具体动作的操作规划，目标是完成当前用户任务。操作步骤应使用英文简洁描述。}
\end{CJK*}

\begin{CJK*}{UTF8}{gbsn}
    \texttt{（3）<|response\_start|> ... <|response\_end|>（可选）：你对用户的语音或文本形式的自然语言回复，内容需简洁、礼貌并与任务状态相关。当当前信息不足以明确执行用户任务时，你可以主动向用户提问以获取更多信息。但请遵循``非必要不提问''的原则，避免频繁打断用户。}
\end{CJK*}

\vspace{1em}

\begin{CJK*}{UTF8}{gbsn}
    \texttt{\# 任务规则}
\end{CJK*}

\begin{CJK*}{UTF8}{gbsn}
    \texttt{在于用户的交互与执行任务的过程中，你需要遵循以下规则：}
\end{CJK*}

\begin{CJK*}{UTF8}{gbsn}
    \texttt{（1）`<|think\_start|> ... <|think\_end|>'和`<|response\_start|> ... <|response\_end|>'中的文本以中文输出，`<|plan\_start|> ... <|plan\_end|>'中的任务规划以英文输出；}
\end{CJK*}

\begin{CJK*}{UTF8}{gbsn}
    \texttt{（2）用户可能会打断你的操作，你需要基于当前画面和用户给你的最新指令决定下一步的操作；}
\end{CJK*}

\begin{CJK*}{UTF8}{gbsn}
    \texttt{（3）`<|plan\_start|>...<|plan\_end|>'中的操作推荐遵循以下格式：}
\end{CJK*}

\hspace{2em}    - put the ... into the ...

\hspace{2em}    - pick up the ...

\hspace{2em}    - navigate to the ...

\hspace{2em}    - open the ...

\hspace{2em}    - close the ...

\begin{CJK*}{UTF8}{gbsn}
    \texttt{（4）`<|plan\_start|>...<|plan\_end|>'中的任务规划一次只能规划下一步的内容，不能规划之后多步的内容，例如：put the A to the B, put the C to the D...这样多个操作不能放到一次规划中，但是它可以出现在你的思考过程中。}
\end{CJK*}

\vspace{1em}

\begin{CJK*}{UTF8}{gbsn}
    \texttt{\# 例子}
\end{CJK*}

User:

\begin{CJK*}{UTF8}{gbsn}
    \texttt{<|system\_info\_start|>你当前看到的画面为...<|system\_info\_end|><|human\_start|>帮我清理桌面。<|human\_end|>}
\end{CJK*}

Assistant:

\begin{CJK*}{UTF8}{gbsn}
    \texttt{<|think\_start|>用户让我清理桌面，当前画面中我能看到.... <|think\_end|><|plan\_start|>put the ... into the ...<|plan\_end|><|response\_start|>收到。<|response\_end|>}
\end{CJK*}

User:

\begin{CJK*}{UTF8}{gbsn}
    \texttt{<|system\_info\_start|>你完成操作后所看到的场景为...<|system\_info\_end|>}
\end{CJK*}

Assistant:

\begin{CJK*}{UTF8}{gbsn}
    \texttt{<|think\_start|>我应该继续收拾桌面，当前我能看到桌面上还有...<|think\_end|><|plan\_start|>put the ... into the ...<|plan\_end|>
}
\end{CJK*}

...
\end{tcolorbox}

\textbf{Reward Model Prompt for RL}
\begin{tcolorbox}
[breakable,colback=white,colframe=black,left=1mm,right=1mm,top=1mm,bottom=1mm,fontupper=\small]
\ttfamily
\# Role

You are a consistency checker for a robot's planning. Your task is to evaluate the robot's thinking process and the resulting final action.

\vspace{1em}

\# Input

1. The robot's thinking process, enclosed by <|think\_start|> and <|think\_end|>.

2. The robot's final action, enclosed by <|plan\_start|> and <|plan\_end|>.

\vspace{1em}

\# Task

You need to determine:

1. Whether the thinking process within <|think\_start|> and <|think\_end|> is reasonable.

2. Whether the final decision within <|plan\_start|> and <|plan\_end|> is consistent with the decision derived from the thinking process in <|think\_start|> and <|think\_end|>.

\vspace{1em}

\# Output

Your output must follow this format: [Your reasoning process] + "The final answer is: [NUMBER]."
\vspace{0.5em}

Here, [NUMBER] is an integer: `1' indicates consistency; `-1' indicates inconsistency; `0' indicates that it is impossible to judge.

\vspace{1em}

\# Notes

1. Ignore ambiguity in names. Items that could potentially be consistent should be considered consistent.

2. Assign -1 only if the thinking process is completely unreasonable.

3. Assign -1 only if it's completely impossible for the thinking process and the plan to be consistent.

4. If the plan is only a part of the strategy outlined in the thinking process, it is still considered consistent (i.e., the plan does not need to reflect all thoughts).
\end{tcolorbox}

\section{Embodied Task-centric Reasoning}
\label{aapx:er}

\subsection{Data Synthesis}

\label{aapx:er_sub}

Publicly available robot datasets often contain long video demonstrations with temporally annotated clips and corresponding individual actions, or they can be segmented automatically into clips using VLMs. To equip \name{} with reasoning and planning capabilities in embodied scenarios, we design three dedicated synthesis pipelines, each targeting a distinct sub-task: task status verification, action affordance, and next action prediction. The details of these pipelines are provided below.

\paragraph{Task Status Verification.} The status of a task is categorized as either complete or incomplete. To construct task verification data labeled as complete, we use the entire video clip as the visual input. Reasoning traces are obtained by prompting a strong VLM (Seed-1.5-VL in thinking mode) with the question ``Is the \{action\} complete?''. We then extract the reasoning content from the response and retain only those instances where the predicted status is complete. Conversely, to represent incomplete task status, we randomly truncate each video clip to between one-half and one-third of its original duration, producing partial visual inputs. As before, we retain only those cases where the predicted status is identified as incomplete.

\paragraph{Action Affordance.} To obtain high-quality labels for action affordance, we distill both labels and reasoning traces from powerful VLMs. However, we find that relying on a single VLM is insufficient for generating accurate labels reliably. Therefore, we employ various VLMs to produce candidate labels and take the intersection of their outputs as the final annotation. The corresponding reasoning traces are extracted from the Seed-1.5-VL-Think model.

\paragraph{Next Action Prediction.} 
The robot datasets are typically annotated with action sequences. Thus, the ground-truth next action can be directly determined once the previous action is specified. To synthesize the corresponding reasoning trace, we leverage Seed-1.5-VL by prompting it with both the video and the following textual input: \textit{``You are an intelligent robot executing \{the task\}. To complete this task, you are performing the action as shown in the video. Your next action is \{next action\}. Please reflect on the rationale behind your next action and output your response from the perspective of a robotic agent.''}

\subsection{Agibot-ER Benchmark}

We construct an embodied task-centric reasoning benchmark based on the Agibot dataset~\cite{bu2025agibot}, which provides manual annotations of tasks, atomic action sequences, and the corresponding video clips for each action. Specifically:
\begin{itemize}
    \item Task Status Verification: We randomly select 100 samples from the synthesized dataset and further verify the label accuracy, resulting in 97 test samples.
    \item Action Affordance: We select 120 video clips and corresponding individual actions from 50 different tasks. To ensure accuracy, we manually annotate these samples rather than relying on VLM-based annotation.
    \item Next Action Prediction. The ground-truth next action is obtained directly from the robot datasets. We randomly sample 400 instances and construct distractor options by leveraging a powerful LLM to generate incorrect candidates that differ subtly from the correct one—such as variations in action type, target object, or destination. To ensure quality, we manually verify all synthesized options, resulting in 381 validated test samples.
\end{itemize}
All test samples are eliminated from the training dataset.

\section{Case studies of RL}\label{appx:case-study}

Our reinforcement learning stage yields several improvements over the SFT model, including: (1) reduced irrational reasoning, (2) improved thought-action consistency, and (3) fewer formatting errors. In the following, we present comparative case studies between \name{}-32B-RL and \name{}-32B-SFT to illustrate the effectiveness of reinforcement learning.

\textbf{Reduced irrational reasoning}
\begin{tcolorbox}
[breakable,colback=white,colframe=black,left=1mm,right=1mm,top=1mm,bottom=1mm,fontupper=\small]
\ttfamily
\# Current task

Put all the drinks on the table into the carton.

\vspace{1em}
\# \name{}-32B-RL output

<|think\_start|>Based on the current scene, I see that there is already a bottle of soda inside the carton, and the remaining items on the table are milk, Oreos, and gum. Recalling my last action, I have already put the coke into the carton. Now, the only beverage left on the table is milk. To complete the task of putting all beverages into the carton, the next step should be to put the milk into the carton. <|think\_end|>

<|plan\_start|>put the milk into the carton<|plan\_end|>

\vspace{1em}
\# \name{}-32B-SFT output

<|think\_start|>Based on the current scene, I see that there is already a bottle of soda inside the carton, and the remaining items on the table are milk, Oreos, and gum. Recalling my last action, I have already put the coke into the carton. Now, all the beverages on the table have been put into the carton, and the task goal has been achieved. Next, I should proactively ask the user if there are any other tasks that need assistance. <|think\_end|>

<|response\_start|>The task has been processed. Is there anything else I can help you with? <|response\_end|>
\end{tcolorbox}

\textbf{Improved thought-action consistency}
\begin{tcolorbox}
[breakable,colback=white,colframe=black,left=1mm,right=1mm,top=1mm,bottom=1mm,fontupper=\small]
\ttfamily
\# Current Task

Put the dishes in the sink

\vspace{1em}
\# \name{}-32B-RL output

<|think\_start|>I have grasped the storage basket with my dual grippers. The basket contains bowls, forks, and spoons. According to the user's instructions, the next step is to take these dishes to the sink. I have now finished collecting the tableware. Next, I should navigate to the sink area and deliver the dishes to the designated location.<|think\_end|>

<|plan\_start|>navigate to the sink<|plan\_end|>

\vspace{1em}
\# \name{}-32B-SFT output

<|think\_start|>I have grasped the storage basket with my dual grippers, ready to transport the dishes. According to the user's instructions, the dishes need to be taken to the sink. All the dishes that need to be washed have been collected in the storage basket. The next step is to transport the storage basket to the sink area, completing the task of collecting and transporting the dishes.<|think\_end|>

<|plan\_start|>navigate to the cupboard<|plan\_end|>

\end{tcolorbox}

\textbf{Fewer formatting errors}

\begin{tcolorbox}
[breakable,colback=white,colframe=black,left=1mm,right=1mm,top=1mm,bottom=1mm,fontupper=\small]
\ttfamily
\# Current Task

Organize the table

\vspace{1em}
\# \name{}-32B-RL output

<|think\_start|>From the current scene, the paper cup is still on the table, near the rubbish bin, and is grasped by my gripper, which indicates that the previous attempt to put the paper cup into the rubbish bin was not successful. Besides the paper cup, there is also a plate with some leftover food residue on the table. To continue organizing the table, I first need to re-attempt putting the paper cup into the rubbish bin to ensure the trash is handled correctly.<|think\_end|>
<|plan\_start|>put the paper cup into the rubbish bin<|plan\_end|>

\vspace{1em}
\# \name{}-32B-SFT output

<|think\_start|>From the current scene, the paper cup is still on the table, near the rubbish bin, and is grasped by my gripper, which indicates that the previous attempt to put the paper cup into the rubbish bin was not successful. Besides the paper cup, there is also a plate with some leftover food residue on the table. To continue organizing the table, I first need to re-attempt putting the paper cup into the rubbish bin to ensure the trash is handled correctly.<|think\_end|>

\begin{verbatim}
{
  "action": "put",
  "object": "paper cup",
  "target": "rubbish bin"
}
\end{verbatim}

\end{tcolorbox}

%% file: paper.bbl
\begin{thebibliography}{88}
\providecommand{\natexlab}[1]{#1}
\providecommand{\url}[1]{\texttt{#1}}
\expandafter\ifx\csname urlstyle\endcsname\relax
  \providecommand{\doi}[1]{doi: #1}\else
  \providecommand{\doi}{doi: \begingroup \urlstyle{rm}\Url}\fi

\bibitem[Ahn et~al.(2022)Ahn, Brohan, Brown, Chebotar, Cortes, David, Finn, Fu, Gopalakrishnan, Hausman, et~al.]{ahn2022can}
Michael Ahn, Anthony Brohan, Noah Brown, Yevgen Chebotar, Omar Cortes, Byron David, Chelsea Finn, Chuyuan Fu, Keerthana Gopalakrishnan, Karol Hausman, et~al.
\newblock Do as i can, not as i say: Grounding language in robotic affordances.
\newblock \emph{arXiv preprint arXiv:2204.01691}, 2022.

\bibitem[Azzolini et~al.(2025)Azzolini, Bai, Brandon, Cao, Chattopadhyay, Chen, Chu, Cui, Diamond, Ding, et~al.]{azzolini2025cosmos}
Alisson Azzolini, Junjie Bai, Hannah Brandon, Jiaxin Cao, Prithvijit Chattopadhyay, Huayu Chen, Jinju Chu, Yin Cui, Jenna Diamond, Yifan Ding, et~al.
\newblock Cosmos-reason1: From physical common sense to embodied reasoning.
\newblock \emph{arXiv preprint arXiv:2503.15558}, 2025.

\bibitem[Bai et~al.(2025)Bai, Chen, Liu, Wang, Ge, Song, Dang, Wang, Wang, Tang, et~al.]{bai2025qwen2}
Shuai Bai, Keqin Chen, Xuejing Liu, Jialin Wang, Wenbin Ge, Sibo Song, Kai Dang, Peng Wang, Shijie Wang, Jun Tang, et~al.
\newblock Qwen2. 5-vl technical report.
\newblock \emph{arXiv preprint arXiv:2502.13923}, 2025.

\bibitem[Baruch et~al.(2021)Baruch, Chen, Dehghan, Feigin, Fu, Gebauer, Kurz, Dimry, Joffe, Schwartz, and Shulman]{baruch2021arkitscenes}
Gilad Baruch, Zhuoyuan Chen, Afshin Dehghan, Yuri Feigin, Peter Fu, Thomas Gebauer, Daniel Kurz, Tal Dimry, Brandon Joffe, Arik Schwartz, and Elad Shulman.
\newblock {ARK}itscenes: A diverse real-world dataset for 3d indoor scene understanding using mobile {RGB}-d data.
\newblock In \emph{Conference on Neural Information Processing Systems Datasets and Benchmarks Track (Round 1)}, 2021.

\bibitem[Belkhale et~al.(2024)Belkhale, Ding, Xiao, Sermanet, Vuong, Tompson, et~al.]{belkhale2024rt}
Suneel Belkhale, Tianli Ding, Ted Xiao, Pierre Sermanet, Quon Vuong, Jonathan Tompson, et~al.
\newblock Rt-h: Action hierarchies using language.
\newblock \emph{arXiv preprint arXiv:2403.01823}, 2024.

\bibitem[Bu et~al.(2025)Bu, Cai, Chen, Cui, Ding, Feng, Gao, He, Hu, Huang, et~al.]{bu2025agibot}
Qingwen Bu, Jisong Cai, Li~Chen, Xiuqi Cui, Yan Ding, Siyuan Feng, Shenyuan Gao, Xindong He, Xuan Hu, Xu~Huang, et~al.
\newblock Agibot world colosseo: A large-scale manipulation platform for scalable and intelligent embodied systems.
\newblock \emph{arXiv preprint arXiv:2503.06669}, 2025.

\bibitem[Cheang et~al.(2025)Cheang, Chen, Cui, Hu, Huang, Kong, Li, Li, Liu, Ma, et~al.]{cheang2025gr}
Chilam Cheang, Sijin Chen, Zhongren Cui, Yingdong Hu, Liqun Huang, Tao Kong, Hang Li, Yifeng Li, Yuxiao Liu, Xiao Ma, et~al.
\newblock Gr-3 technical report.
\newblock \emph{arXiv preprint arXiv:2507.15493}, 2025.

\bibitem[Cheng et~al.(2024)Cheng, Ji, Yang, Gongye, Zou, Kautz, B{\i}y{\i}k, Yin, Liu, and Wang]{cheng2024navila}
An-Chieh Cheng, Yandong Ji, Zhaojing Yang, Zaitian Gongye, Xueyan Zou, Jan Kautz, Erdem B{\i}y{\i}k, Hongxu Yin, Sifei Liu, and Xiaolong Wang.
\newblock Navila: Legged robot vision-language-action model for navigation.
\newblock \emph{arXiv preprint arXiv:2412.04453}, 2024.

\bibitem[Cheng et~al.(2025)Cheng, Zhang, Zhang, Yang, Guan, Wu, Li, Zhang, Liu, Mai, et~al.]{cheng2025simplevqa}
Xianfu Cheng, Wei Zhang, Shiwei Zhang, Jian Yang, Xiangyuan Guan, Xianjie Wu, Xiang Li, Ge~Zhang, Jiaheng Liu, Yuying Mai, et~al.
\newblock Simplevqa: Multimodal factuality evaluation for multimodal large language models.
\newblock \emph{arXiv preprint arXiv:2502.13059}, 2025.

\bibitem[Chi et~al.(2024)Chi, Xu, Pan, Cousineau, Burchfiel, Feng, Tedrake, and Song]{chi2024universal}
Cheng Chi, Zhenjia Xu, Chuer Pan, Eric Cousineau, Benjamin Burchfiel, Siyuan Feng, Russ Tedrake, and Shuran Song.
\newblock Universal manipulation interface: In-the-wild robot teaching without in-the-wild robots.
\newblock In \emph{Proceedings of Robotics: Science and Systems (RSS)}, 2024.

\bibitem[Colombani et~al.(2024)Colombani, Ognibene, and Boccignone]{colombani2024one}
Simone Colombani, Dimitri Ognibene, and Giuseppe Boccignone.
\newblock One to rule them all: natural language to bind communication, perception and action.
\newblock \emph{arXiv preprint arXiv:2411.15033}, 2024.

\bibitem[Dai et~al.(2017)Dai, Chang, Savva, Halber, Funkhouser, and Nie{\ss}ner]{dai2017scannet}
Angela Dai, Angel~X Chang, Manolis Savva, Maciej Halber, Thomas Funkhouser, and Matthias Nie{\ss}ner.
\newblock Scannet: Richly-annotated 3d reconstructions of indoor scenes.
\newblock In \emph{Proceedings of the IEEE conference on computer vision and pattern recognition}, pages 5828--5839, 2017.

\bibitem[Dai et~al.(2024)Dai, Lee, Fazeli, and Chai]{dai2024racer}
Yinpei Dai, Jayjun Lee, Nima Fazeli, and Joyce Chai.
\newblock Racer: Rich language-guided failure recovery policies for imitation learning.
\newblock \emph{arXiv preprint arXiv:2409.14674}, 2024.

\bibitem[Deitke et~al.(2024)Deitke, Clark, Lee, Tripathi, Yang, Park, Salehi, Muennighoff, Lo, Soldaini, et~al.]{deitke2024molmo}
Matt Deitke, Christopher Clark, Sangho Lee, Rohun Tripathi, Yue Yang, Jae~Sung Park, Mohammadreza Salehi, Niklas Muennighoff, Kyle Lo, Luca Soldaini, et~al.
\newblock Molmo and pixmo: Open weights and open data for state-of-the-art multimodal models.
\newblock \emph{arXiv e-prints}, pages arXiv--2409, 2024.

\bibitem[Driess et~al.(2023)Driess, Xia, Sajjadi, Lynch, Chowdhery, Wahid, Tompson, Vuong, Yu, et~al.]{driess2023palm}
Danny Driess, Fei Xia, Mehdi~SM Sajjadi, Corey Lynch, Aakanksha Chowdhery, Ayzaan Wahid, Jonathan Tompson, Quan Vuong, Tianhe Yu, et~al.
\newblock Palm-e: An embodied multimodal language model.
\newblock 2023.

\bibitem[Du et~al.(2024)Du, Wu, Li, Huang, and Wei]{du2024embspatial}
Mengfei Du, Binhao Wu, Zejun Li, Xuanjing Huang, and Zhongyu Wei.
\newblock Embspatial-bench: Benchmarking spatial understanding for embodied tasks with large vision-language models.
\newblock \emph{Annual Meeting of the Association for Computational Linguistics (Volume 2: Short Papers)}, 2024.

\bibitem[Duan et~al.(2024)Duan, Yuan, Pumacay, Wang, Ehsani, Fox, and Krishna]{duan2024manipulate}
Jiafei Duan, Wentao Yuan, Wilbert Pumacay, Yi~Ru Wang, Kiana Ehsani, Dieter Fox, and Ranjay Krishna.
\newblock Manipulate-anything: Automating real-world robots using vision-language models.
\newblock \emph{arXiv preprint arXiv:2406.18915}, 2024.

\bibitem[Fu et~al.(2024{\natexlab{a}})Fu, Zhang, Yin, Li, Fang, Zhao, Duan, Sun, Liu, Wang, et~al.]{fu2024mme}
Chaoyou Fu, Yi-Fan Zhang, Shukang Yin, Bo~Li, Xinyu Fang, Sirui Zhao, Haodong Duan, Xing Sun, Ziwei Liu, Liang Wang, et~al.
\newblock Mme-survey: A comprehensive survey on evaluation of multimodal llms.
\newblock \emph{arXiv preprint arXiv:2411.15296}, 2024{\natexlab{a}}.

\bibitem[Fu et~al.(2025)Fu, Dai, Luo, Li, Ren, Zhang, Wang, Zhou, Shen, Zhang, et~al.]{fu2025videomme}
Chaoyou Fu, Yuhan Dai, Yongdong Luo, Lei Li, Shuhuai Ren, Renrui Zhang, Zihan Wang, Chenyu Zhou, Yunhang Shen, Mengdan Zhang, et~al.
\newblock Video-mme: The first-ever comprehensive evaluation benchmark of multi-modal llms in video analysis.
\newblock In \emph{Proceedings of the Computer Vision and Pattern Recognition Conference}, pages 24108--24118, 2025.

\bibitem[Fu et~al.(2024{\natexlab{b}})Fu, Hu, Li, Feng, Wang, Lin, Roth, Smith, Ma, and Krishna]{fu2024blink}
Xingyu Fu, Yushi Hu, Bangzheng Li, Yu~Feng, Haoyu Wang, Xudong Lin, Dan Roth, Noah~A Smith, Wei-Chiu Ma, and Ranjay Krishna.
\newblock Blink: Multimodal large language models can see but not perceive.
\newblock In \emph{European Conference on Computer Vision}, pages 148--166. Springer, 2024{\natexlab{b}}.

\bibitem[Gong et~al.(2025)Gong, Hou, Li, Li, Lian, Liu, Liu, Liu, Lu, Shi, Sun, Tian, Tian, Wang, Wang, Wang, Wu, Wu, Xia, Xiao, Yang, Zhai, Zhang, Zhang, Zhang, Zhao, Yang, and Huang]{gong2025seedream20nativechineseenglish}
Lixue Gong, Xiaoxia Hou, Fanshi Li, Liang Li, Xiaochen Lian, Fei Liu, Liyang Liu, Wei Liu, Wei Lu, Yichun Shi, Shiqi Sun, Yu~Tian, Zhi Tian, Peng Wang, Xun Wang, Ye~Wang, Guofeng Wu, Jie Wu, Xin Xia, Xuefeng Xiao, Linjie Yang, Zhonghua Zhai, Xinyu Zhang, Qi~Zhang, Yuwei Zhang, Shijia Zhao, Jianchao Yang, and Weilin Huang.
\newblock Seedream 2.0: A native chinese-english bilingual image generation foundation model, 2025.

\bibitem[Grauman et~al.(2022)Grauman, Westbury, Byrne, Chavis, Furnari, Girdhar, Hamburger, Jiang, Liu, Liu, et~al.]{grauman2022ego4d}
Kristen Grauman, Andrew Westbury, Eugene Byrne, Zachary Chavis, Antonino Furnari, Rohit Girdhar, Jackson Hamburger, Hao Jiang, Miao Liu, Xingyu Liu, et~al.
\newblock Ego4d: Around the world in 3,000 hours of egocentric video.
\newblock In \emph{Proceedings of the IEEE/CVF conference on computer vision and pattern recognition}, pages 18995--19012, 2022.

\bibitem[Guo et~al.(2025{\natexlab{a}})Guo, Yang, Zhang, Song, Zhang, Xu, Zhu, Ma, Wang, Bi, et~al.]{guo2025deepseek}
Daya Guo, Dejian Yang, Haowei Zhang, Junxiao Song, Ruoyu Zhang, Runxin Xu, Qihao Zhu, Shirong Ma, Peiyi Wang, Xiao Bi, et~al.
\newblock Deepseek-r1: Incentivizing reasoning capability in llms via reinforcement learning.
\newblock \emph{arXiv preprint arXiv:2501.12948}, 2025{\natexlab{a}}.

\bibitem[Guo et~al.(2025{\natexlab{b}})Guo, Wu, Zhu, Leng, Shi, Chen, Fan, Wang, Jiang, Wang, et~al.]{guo2025seed1}
Dong Guo, Faming Wu, Feida Zhu, Fuxing Leng, Guang Shi, Haobin Chen, Haoqi Fan, Jian Wang, Jianyu Jiang, Jiawei Wang, et~al.
\newblock Seed1. 5-vl technical report.
\newblock \emph{arXiv preprint arXiv:2505.07062}, 2025{\natexlab{b}}.

\bibitem[Gupta et~al.(2019)Gupta, Dollar, and Girshick]{gupta2019lvis}
Agrim Gupta, Piotr Dollar, and Ross Girshick.
\newblock Lvis: A dataset for large vocabulary instance segmentation.
\newblock In \emph{Proceedings of the IEEE/CVF conference on computer vision and pattern recognition}, pages 5356--5364, 2019.

\bibitem[Hong et~al.(2025)Hong, Yu, Gu, Wang, Gan, Tang, Cheng, Qi, Ji, Pan, et~al.]{hong2025glm}
Wenyi Hong, Wenmeng Yu, Xiaotao Gu, Guo Wang, Guobing Gan, Haomiao Tang, Jiale Cheng, Ji~Qi, Junhui Ji, Lihang Pan, et~al.
\newblock Glm-4.1 v-thinking: Towards versatile multimodal reasoning with scalable reinforcement learning.
\newblock \emph{arXiv preprint arXiv:2507.01006}, 2025.

\bibitem[Hoque et~al.(2025)Hoque, Huang, Yoon, Sivapurapu, and Zhang]{hoque2025egodex}
Ryan Hoque, Peide Huang, David~J Yoon, Mouli Sivapurapu, and Jian Zhang.
\newblock Egodex: Learning dexterous manipulation from large-scale egocentric video.
\newblock \emph{arXiv preprint arXiv:2505.11709}, 2025.

\bibitem[Hu et~al.(2023)Hu, Lin, Zhang, Yi, and Gao]{hu2023look}
Yingdong Hu, Fanqi Lin, Tong Zhang, Li~Yi, and Yang Gao.
\newblock Look before you leap: Unveiling the power of gpt-4v in robotic vision-language planning.
\newblock \emph{arXiv preprint arXiv:2311.17842}, 2023.

\bibitem[Huang et~al.(2022)Huang, Xia, Xiao, Chan, Liang, Florence, Zeng, Tompson, Mordatch, Chebotar, et~al.]{huang2022inner}
Wenlong Huang, Fei Xia, Ted Xiao, Harris Chan, Jacky Liang, Pete Florence, Andy Zeng, Jonathan Tompson, Igor Mordatch, Yevgen Chebotar, et~al.
\newblock Inner monologue: Embodied reasoning through planning with language models.
\newblock \emph{arXiv preprint arXiv:2207.05608}, 2022.

\bibitem[Hurst et~al.(2024)Hurst, Lerer, Goucher, Perelman, Ramesh, Clark, Ostrow, Welihinda, Hayes, Radford, et~al.]{hurst2024gpt}
Aaron Hurst, Adam Lerer, Adam~P Goucher, Adam Perelman, Aditya Ramesh, Aidan Clark, AJ~Ostrow, Akila Welihinda, Alan Hayes, Alec Radford, et~al.
\newblock Gpt-4o system card.
\newblock \emph{arXiv preprint arXiv:2410.21276}, 2024.

\bibitem[Jia et~al.(2022)Jia, Lei, Zhu, and Huang]{jia2022egotaskqa}
Baoxiong Jia, Ting Lei, Song-Chun Zhu, and Siyuan Huang.
\newblock Egotaskqa: Understanding human tasks in egocentric videos.
\newblock \emph{Advances in Neural Information Processing Systems}, 35:\penalty0 3343--3360, 2022.

\bibitem[Khazatsky et~al.(2024)Khazatsky, Pertsch, Nair, Balakrishna, Dasari, Karamcheti, Nasiriany, Srirama, Chen, Ellis, et~al.]{khazatsky2024droid}
Alexander Khazatsky, Karl Pertsch, Suraj Nair, Ashwin Balakrishna, Sudeep Dasari, Siddharth Karamcheti, Soroush Nasiriany, Mohan~Kumar Srirama, Lawrence~Yunliang Chen, Kirsty Ellis, et~al.
\newblock Droid: A large-scale in-the-wild robot manipulation dataset.
\newblock \emph{arXiv preprint arXiv:2403.12945}, 2024.

\bibitem[Kingma and Ba(2015)]{2015-kingma}
Diederik~P. Kingma and Jimmy Ba.
\newblock Adam: A method for stochastic optimization.
\newblock In \emph{International Conference on Learning Representations}, 2015.

\bibitem[Kirillova et~al.(2021)Kirillova, Lyapustin, Antsiferova, and Vatolin]{kirillova2021erqa}
Anastasia Kirillova, Eugene Lyapustin, Anastasia Antsiferova, and Dmitry Vatolin.
\newblock Erqa: Edge-restoration quality assessment for video super-resolution.
\newblock \emph{arXiv preprint arXiv:2110.09992}, 2021.

\bibitem[Lazarow et~al.(2025)Lazarow, Griffiths, Kohavi, Crespo, and Dehghan]{lazarow2025cubify}
Justin Lazarow, David Griffiths, Gefen Kohavi, Francisco Crespo, and Afshin Dehghan.
\newblock Cubify anything: Scaling indoor 3d object detection.
\newblock In \emph{Proceedings of the Computer Vision and Pattern Recognition Conference}, pages 22225--22233, 2025.

\bibitem[Lin et~al.(2025)Lin, Nie, Zai, Wei, Han, Xu, Niu, Han, Lin, Lu, et~al.]{lin2025evolvenav}
Bingqian Lin, Yunshuang Nie, Khun~Loun Zai, Ziming Wei, Mingfei Han, Rongtao Xu, Minzhe Niu, Jianhua Han, Liang Lin, Cewu Lu, et~al.
\newblock Evolvenav: Self-improving embodied reasoning for llm-based vision-language navigation.
\newblock \emph{arXiv preprint arXiv:2506.01551}, 2025.

\bibitem[Liu et~al.(2023{\natexlab{a}})Liu, Emerson, and Collier]{liu2023vsr}
Fangyu Liu, Guy Emerson, and Nigel Collier.
\newblock Visual spatial reasoning.
\newblock \emph{Transactions of the Association for Computational Linguistics}, 11:\penalty0 635--651, 2023{\natexlab{a}}.

\bibitem[Liu et~al.(2023{\natexlab{b}})Liu, Chen, Zhu, Swaminathan, Kolobov, and Cheng]{liu2023interactive}
Huihan Liu, Alice Chen, Yuke Zhu, Adith Swaminathan, Andrey Kolobov, and Ching-An Cheng.
\newblock Interactive robot learning from verbal correction.
\newblock \emph{arXiv preprint arXiv:2310.17555}, 2023{\natexlab{b}}.

\bibitem[Liu et~al.(2024{\natexlab{a}})Liu, Song, Lin, Lam, Neubig, Li, and Yue]{liu2024visualwebbench}
Junpeng Liu, Yifan Song, Bill~Yuchen Lin, Wai Lam, Graham Neubig, Yuanzhi Li, and Xiang Yue.
\newblock Visualwebbench: How far have multimodal llms evolved in web page understanding and grounding?
\newblock \emph{arXiv preprint arXiv:2404.05955}, 2024{\natexlab{a}}.

\bibitem[Liu et~al.(2024{\natexlab{b}})Liu, Duan, Zhang, Li, Zhang, Zhao, Yuan, Wang, He, Liu, et~al.]{liu2024mmbench}
Yuan Liu, Haodong Duan, Yuanhan Zhang, Bo~Li, Songyang Zhang, Wangbo Zhao, Yike Yuan, Jiaqi Wang, Conghui He, Ziwei Liu, et~al.
\newblock Mmbench: Is your multi-modal model an all-around player?
\newblock In \emph{European conference on computer vision}, pages 216--233. Springer, 2024{\natexlab{b}}.

\bibitem[Loshchilov and Hutter(2019)]{loshchilov2018decoupled}
Ilya Loshchilov and Frank Hutter.
\newblock Decoupled weight decay regularization.
\newblock In \emph{International Conference on Learning Representations}, 2019.

\bibitem[Lu et~al.(2023)Lu, Bansal, Xia, Liu, Li, Hajishirzi, Cheng, Chang, Galley, and Gao]{lu2023mathvista}
Pan Lu, Hritik Bansal, Tony Xia, Jiacheng Liu, Chunyuan Li, Hannaneh Hajishirzi, Hao Cheng, Kai-Wei Chang, Michel Galley, and Jianfeng Gao.
\newblock Mathvista: Evaluating mathematical reasoning of foundation models in visual contexts.
\newblock \emph{arXiv preprint arXiv:2310.02255}, 2023.

\bibitem[Majumdar et~al.(2024)Majumdar, Ajay, Zhang, Putta, Yenamandra, Henaff, Silwal, Mcvay, Maksymets, Arnaud, et~al.]{majumdar2024openeqa}
Arjun Majumdar, Anurag Ajay, Xiaohan Zhang, Pranav Putta, Sriram Yenamandra, Mikael Henaff, Sneha Silwal, Paul Mcvay, Oleksandr Maksymets, Sergio Arnaud, et~al.
\newblock Openeqa: Embodied question answering in the era of foundation models.
\newblock In \emph{Proceedings of the IEEE/CVF conference on computer vision and pattern recognition}, pages 16488--16498, 2024.

\bibitem[Mangalam et~al.(2023)Mangalam, Akshulakov, and Malik]{mangalam2023egoschema}
Karttikeya Mangalam, Raiymbek Akshulakov, and Jitendra Malik.
\newblock Egoschema: A diagnostic benchmark for very long-form video language understanding.
\newblock \emph{Advances in Neural Information Processing Systems}, 36:\penalty0 46212--46244, 2023.

\bibitem[Matuszek et~al.(2013)Matuszek, Herbst, Zettlemoyer, and Fox]{matuszek2013learning}
Cynthia Matuszek, Evan Herbst, Luke Zettlemoyer, and Dieter Fox.
\newblock Learning to parse natural language commands to a robot control system.
\newblock In \emph{Experimental robotics: the 13th international symposium on experimental robotics}, pages 403--415. Springer, 2013.

\bibitem[McCallum et~al.(2023)McCallum, Taylor-Davies, Albrecht, and Suglia]{mccallum2023feedback}
Sabrina McCallum, Max Taylor-Davies, Stefano Albrecht, and Alessandro Suglia.
\newblock Is feedback all you need? leveraging natural language feedback in goal-conditioned rl.
\newblock In \emph{NeurIPS 2023 Workshop on Goal-Conditioned Reinforcement Learning}, 2023.

\bibitem[Mei et~al.(2024)Mei, Zhu, Zhang, and Gan]{mei2024replanvlm}
Aoran Mei, Guo-Niu Zhu, Huaxiang Zhang, and Zhongxue Gan.
\newblock Replanvlm: Replanning robotic tasks with visual language models.
\newblock \emph{IEEE Robotics and Automation Letters}, 2024.

\bibitem[Namasivayam et~al.(2023)Namasivayam, Singh, Bindal, Tuli, Agrawal, Jain, Singla, and Paul]{namasivayam2023learning}
K~Namasivayam, Himanshu Singh, Vishal Bindal, Arnav Tuli, Vishwajeet Agrawal, Rahul Jain, Parag Singla, and Rohan Paul.
\newblock Learning neuro-symbolic programs for language guided robot manipulation.
\newblock In \emph{2023 IEEE International Conference on Robotics and Automation (ICRA)}, pages 7973--7980. IEEE, 2023.

\bibitem[Patki et~al.(2019)Patki, Daniele, Walter, and Howard]{patki2019inferring}
Siddharth Patki, Andrea~F Daniele, Matthew~R Walter, and Thomas~M Howard.
\newblock Inferring compact representations for efficient natural language understanding of robot instructions.
\newblock In \emph{2019 International Conference on Robotics and Automation (ICRA)}, pages 6926--6933. IEEE, 2019.

\bibitem[Qin et~al.(2025)Qin, Ye, Fang, Wang, Liang, Tian, Zhang, Li, Li, Huang, et~al.]{qin2025ui}
Yujia Qin, Yining Ye, Junjie Fang, Haoming Wang, Shihao Liang, Shizuo Tian, Junda Zhang, Jiahao Li, Yunxin Li, Shijue Huang, et~al.
\newblock Ui-tars: Pioneering automated gui interaction with native agents.
\newblock \emph{arXiv preprint arXiv:2501.12326}, 2025.

\bibitem[Ray et~al.(2024)Ray, Duan, Brown, Tan, Bashkirova, Hendrix, Ehsani, Kembhavi, Plummer, Krishna, et~al.]{ray2024sat}
Arijit Ray, Jiafei Duan, Ellis Brown, Reuben Tan, Dina Bashkirova, Rose Hendrix, Kiana Ehsani, Aniruddha Kembhavi, Bryan~A Plummer, Ranjay Krishna, et~al.
\newblock Sat: Dynamic spatial aptitude training for multimodal language models.
\newblock \emph{arXiv preprint arXiv:2412.07755}, 2024.

\bibitem[Sermanet et~al.(2024)Sermanet, Ding, Zhao, Xia, Dwibedi, Gopalakrishnan, Chan, Dulac-Arnold, Maddineni, Joshi, et~al.]{sermanet2024robovqa}
Pierre Sermanet, Tianli Ding, Jeffrey Zhao, Fei Xia, Debidatta Dwibedi, Keerthana Gopalakrishnan, Christine Chan, Gabriel Dulac-Arnold, Sharath Maddineni, Nikhil~J Joshi, et~al.
\newblock Robovqa: Multimodal long-horizon reasoning for robotics.
\newblock In \emph{2024 IEEE International Conference on Robotics and Automation (ICRA)}, pages 645--652. IEEE, 2024.

\bibitem[Shah et~al.(2024)Shah, Yu, Zhu, Zhu, and Mart{\'\i}n-Mart{\'\i}n]{shah2024bumble}
Rutav Shah, Albert Yu, Yifeng Zhu, Yuke Zhu, and Roberto Mart{\'\i}n-Mart{\'\i}n.
\newblock Bumble: Unifying reasoning and acting with vision-language models for building-wide mobile manipulation.
\newblock \emph{arXiv preprint arXiv:2410.06237}, 2024.

\bibitem[Shao et~al.(2024)Shao, Wang, Zhu, Xu, Song, Bi, Zhang, Zhang, Li, Wu, et~al.]{shao2024deepseekmath}
Zhihong Shao, Peiyi Wang, Qihao Zhu, Runxin Xu, Junxiao Song, Xiao Bi, Haowei Zhang, Mingchuan Zhang, YK~Li, Yang Wu, et~al.
\newblock Deepseekmath: Pushing the limits of mathematical reasoning in open language models.
\newblock \emph{arXiv preprint arXiv:2402.03300}, 2024.

\bibitem[Sheng et~al.(2024)Sheng, Zhang, Ye, Wu, Zhang, Zhang, Peng, Lin, and Wu]{sheng2024hybridflow}
Guangming Sheng, Chi Zhang, Zilingfeng Ye, Xibin Wu, Wang Zhang, Ru~Zhang, Yanghua Peng, Haibin Lin, and Chuan Wu.
\newblock Hybridflow: A flexible and efficient rlhf framework.
\newblock \emph{arXiv preprint arXiv: 2409.19256}, 2024.

\bibitem[Shi et~al.(2024)Shi, Hu, Zhao, Sharma, Pertsch, et~al.]{shi2024yell}
Lucy~Xiaoyang Shi, Zheyuan Hu, Tony~Z Zhao, Archit Sharma, Karl Pertsch, et~al.
\newblock Yell at your robot: Improving on-the-fly from language corrections.
\newblock \emph{arXiv preprint arXiv:2403.12910}, 2024.

\bibitem[Shi et~al.(2025)Shi, brian ichter, Equi, Ke, Pertsch, Vuong, Tanner, Walling, Wang, Fusai, Li-Bell, Driess, Groom, Levine, and Finn]{shi2025hi}
Lucy~Xiaoyang Shi, brian ichter, Michael~Robert Equi, Liyiming Ke, Karl Pertsch, Quan Vuong, James Tanner, Anna Walling, Haohuan Wang, Niccolo Fusai, Adrian Li-Bell, Danny Driess, Lachy Groom, Sergey Levine, and Chelsea Finn.
\newblock Hi robot: Open-ended instruction following with hierarchical vision-language-action models.
\newblock In \emph{International Conference on Machine Learning}, 2025.

\bibitem[Singh et~al.(2024)Singh, Bhattacharyya, and Namboodiri]{singh2024lgr2}
Utsav Singh, Pramit Bhattacharyya, and Vinay~P Namboodiri.
\newblock Lgr2: Language guided reward relabeling for accelerating hierarchical reinforcement learning.
\newblock \emph{arXiv preprint arXiv:2406.05881}, 2024.

\bibitem[Skreta et~al.(2024)Skreta, Zhou, Yuan, Darvish, Aspuru-Guzik, and Garg]{skreta2024replan}
Marta Skreta, Zihan Zhou, Jia~Lin Yuan, Kourosh Darvish, Al{\'a}n Aspuru-Guzik, and Animesh Garg.
\newblock Replan: Robotic replanning with perception and language models.
\newblock \emph{arXiv preprint arXiv:2401.04157}, 2024.

\bibitem[Song et~al.(2023)Song, Wu, Washington, Sadler, Chao, and Su]{song2023llm}
Chan~Hee Song, Jiaman Wu, Clayton Washington, Brian~M Sadler, Wei-Lun Chao, and Yu~Su.
\newblock Llm-planner: Few-shot grounded planning for embodied agents with large language models.
\newblock In \emph{Proceedings of the IEEE/CVF international conference on computer vision}, pages 2998--3009, 2023.

\bibitem[Song et~al.(2015)Song, Lichtenberg, and Xiao]{song2015sun}
Shuran Song, Samuel~P Lichtenberg, and Jianxiong Xiao.
\newblock Sun rgb-d: A rgb-d scene understanding benchmark suite.
\newblock In \emph{Proceedings of the IEEE conference on computer vision and pattern recognition}, pages 567--576, 2015.

\bibitem[Swadzba et~al.(2009)Swadzba, Wachsmuth, Vorwerg, and Rickheit]{swadzba2009computational}
Agnes Swadzba, Sven Wachsmuth, Constanze Vorwerg, and Gert Rickheit.
\newblock A computational model for the alignment of hierarchical scene representations in human-robot interaction.
\newblock In \emph{IJCAI}, pages 1857--1863, 2009.

\bibitem[Team et~al.(2025{\natexlab{a}})Team, Cao, Tan, Ji, Lin, Li, Cao, Wang, Zhou, Han, et~al.]{team2025robobrain}
BAAI~RoboBrain Team, Mingyu Cao, Huajie Tan, Yuheng Ji, Minglan Lin, Zhiyu Li, Zhou Cao, Pengwei Wang, Enshen Zhou, Yi~Han, et~al.
\newblock Robobrain 2.0 technical report.
\newblock \emph{arXiv preprint arXiv:2507.02029}, 2025{\natexlab{a}}.

\bibitem[Team et~al.(2025{\natexlab{b}})Team, Abeyruwan, Ainslie, Alayrac, Arenas, Armstrong, Balakrishna, Baruch, Bauza, Blokzijl, et~al.]{team2025gemini}
Gemini~Robotics Team, Saminda Abeyruwan, Joshua Ainslie, Jean-Baptiste Alayrac, Montserrat~Gonzalez Arenas, Travis Armstrong, Ashwin Balakrishna, Robert Baruch, Maria Bauza, Michiel Blokzijl, et~al.
\newblock Gemini robotics: Bringing ai into the physical world.
\newblock \emph{arXiv preprint arXiv:2503.20020}, 2025{\natexlab{b}}.

\bibitem[Tong et~al.(2024)Tong, Brown, Wu, Woo, IYER, Akula, Yang, Yang, Middepogu, Wang, et~al.]{tong2024cambrian}
Peter Tong, Ellis Brown, Penghao Wu, Sanghyun Woo, Adithya Jairam~Vedagiri IYER, Sai~Charitha Akula, Shusheng Yang, Jihan Yang, Manoj Middepogu, Ziteng Wang, et~al.
\newblock Cambrian-1: A fully open, vision-centric exploration of multimodal llms.
\newblock \emph{Advances in Neural Information Processing Systems}, 37:\penalty0 87310--87356, 2024.

\bibitem[Wald et~al.(2019)Wald, Avetisyan, Navab, Tombari, and Nie{\ss}ner]{wald2019rio}
Johanna Wald, Armen Avetisyan, Nassir Navab, Federico Tombari, and Matthias Nie{\ss}ner.
\newblock Rio: 3d object instance re-localization in changing indoor environments.
\newblock In \emph{Proceedings of the IEEE/CVF International Conference on Computer Vision}, pages 7658--7667, 2019.

\bibitem[Walke et~al.(2023)Walke, Black, Zhao, Vuong, Zheng, Hansen-Estruch, He, Myers, Kim, Du, et~al.]{walke2023bridgedata}
Homer~Rich Walke, Kevin Black, Tony~Z Zhao, Quan Vuong, Chongyi Zheng, Philippe Hansen-Estruch, Andre~Wang He, Vivek Myers, Moo~Jin Kim, Max Du, et~al.
\newblock Bridgedata v2: A dataset for robot learning at scale.
\newblock In \emph{Conference on Robot Learning}, pages 1723--1736. PMLR, 2023.

\bibitem[Wang et~al.(2024{\natexlab{a}})Wang, Zhang, Dong, Fang, and Feng]{wang2024vlm}
Beichen Wang, Juexiao Zhang, Shuwen Dong, Irving Fang, and Chen Feng.
\newblock Vlm see, robot do: Human demo video to robot action plan via vision language model.
\newblock \emph{arXiv preprint arXiv:2410.08792}, 2024{\natexlab{a}}.

\bibitem[Wang et~al.(2024{\natexlab{b}})Wang, Pan, Shi, Lu, Ren, Zhou, Zhan, and Li]{wang2024measuring}
Ke~Wang, Junting Pan, Weikang Shi, Zimu Lu, Houxing Ren, Aojun Zhou, Mingjie Zhan, and Hongsheng Li.
\newblock Measuring multimodal mathematical reasoning with math-vision dataset.
\newblock \emph{Advances in Neural Information Processing Systems}, 37:\penalty0 95095--95169, 2024{\natexlab{b}}.

\bibitem[Wang et~al.(2023)Wang, Kwon, Rad, Pan, Chakraborty, Andrist, Bohus, Feniello, Tekin, Frujeri, et~al.]{wang2023holoassist}
Xin Wang, Taein Kwon, Mahdi Rad, Bowen Pan, Ishani Chakraborty, Sean Andrist, Dan Bohus, Ashley Feniello, Bugra Tekin, Felipe~Vieira Frujeri, et~al.
\newblock Holoassist: an egocentric human interaction dataset for interactive ai assistants in the real world.
\newblock In \emph{Proceedings of the IEEE/CVF International Conference on Computer Vision}, pages 20270--20281, 2023.

\bibitem[xAI(2024)]{realworldqa}
xAI.
\newblock Realworldqa: A benchmark for real-world spatial understanding.
\newblock 2024.
\newblock URL \url{https://huggingface.co/datasets/xai-org/RealworldQA}.

\bibitem[Xiao et~al.(2024)Xiao, Janaka, Hu, Gupta, Li, Yu, and Hsu]{xiao2024robi}
Anxing Xiao, Nuwan Janaka, Tianrun Hu, Anshul Gupta, Kaixin Li, Cunjun Yu, and David Hsu.
\newblock Robi butler: Remote multimodal interactions with household robot assistant.
\newblock \emph{arXiv e-prints}, pages arXiv--2409, 2024.

\bibitem[Xiao et~al.(2021)Xiao, Shang, Yao, and Chua]{xiao2021nextqa}
Junbin Xiao, Xindi Shang, Angela Yao, and Tat-Seng Chua.
\newblock Next-qa: Next phase of question-answering to explaining temporal actions.
\newblock In \emph{Proceedings of the IEEE/CVF conference on computer vision and pattern recognition}, pages 9777--9786, 2021.

\bibitem[Yang et~al.(2024{\natexlab{a}})Yang, Yang, Zhang, Hui, Zheng, Yu, Li, Liu, Huang, Wei, et~al.]{yang2024qwen2}
An~Yang, Baosong Yang, Beichen Zhang, Binyuan Hui, Bo~Zheng, Bowen Yu, Chengyuan Li, Dayiheng Liu, Fei Huang, Haoran Wei, et~al.
\newblock Qwen2. 5 technical report.
\newblock \emph{arXiv preprint arXiv:2412.15115}, 2024{\natexlab{a}}.

\bibitem[Yang et~al.(2025)Yang, Yang, Gupta, Han, Fei-Fei, and Xie]{yang2025thinking}
Jihan Yang, Shusheng Yang, Anjali~W Gupta, Rilyn Han, Li~Fei-Fei, and Saining Xie.
\newblock Thinking in space: How multimodal large language models see, remember, and recall spaces.
\newblock In \emph{Proceedings of the Computer Vision and Pattern Recognition Conference}, pages 10632--10643, 2025.

\bibitem[Yang et~al.(2024{\natexlab{b}})Yang, Kang, Huang, Zhao, Xu, Feng, and Zhao]{yang2024depth}
Lihe Yang, Bingyi Kang, Zilong Huang, Zhen Zhao, Xiaogang Xu, Jiashi Feng, and Hengshuang Zhao.
\newblock Depth anything v2.
\newblock \emph{Advances in Neural Information Processing Systems}, 37:\penalty0 21875--21911, 2024{\natexlab{b}}.

\bibitem[Yang et~al.(2024{\natexlab{c}})Yang, Garrett, Fox, Lozano-P{\'e}rez, and Kaelbling]{yang2024guiding}
Zhutian Yang, Caelan Garrett, Dieter Fox, Tom{\'a}s Lozano-P{\'e}rez, and Leslie~Pack Kaelbling.
\newblock Guiding long-horizon task and motion planning with vision language models.
\newblock \emph{arXiv preprint arXiv:2410.02193}, 2024{\natexlab{c}}.

\bibitem[Yang et~al.(2024{\natexlab{d}})Yang, Li, Yan, Zhang, Huang, and Liu]{yang2024react}
Zonghan Yang, Peng Li, Ming Yan, Ji~Zhang, Fei Huang, and Yang Liu.
\newblock React meets actre: Autonomous annotation of agent trajectories for contrastive self-training.
\newblock In \emph{First Conference on Language Modeling}, 2024{\natexlab{d}}.

\bibitem[Yeshwanth et~al.(2023)Yeshwanth, Liu, Nie{\ss}ner, and Dai]{yeshwanth2023scannet++}
Chandan Yeshwanth, Yueh-Cheng Liu, Matthias Nie{\ss}ner, and Angela Dai.
\newblock Scannet++: A high-fidelity dataset of 3d indoor scenes.
\newblock In \emph{Proceedings of the IEEE/CVF International Conference on Computer Vision}, 2023.

\bibitem[Yu et~al.(2016)Yu, Poirson, Yang, Berg, and Berg]{yu2016modeling}
Licheng Yu, Patrick Poirson, Shan Yang, Alexander~C Berg, and Tamara~L Berg.
\newblock Modeling context in referring expressions.
\newblock In \emph{European conference on computer vision}, pages 69--85. Springer, 2016.

\bibitem[Yuan et~al.(2024)Yuan, Duan, Blukis, Pumacay, Krishna, Murali, Mousavian, and Fox]{yuan2024robopoint}
Wentao Yuan, Jiafei Duan, Valts Blukis, Wilbert Pumacay, Ranjay Krishna, Adithyavairavan Murali, Arsalan Mousavian, and Dieter Fox.
\newblock Robopoint: A vision-language model for spatial affordance prediction for robotics.
\newblock \emph{arXiv preprint arXiv:2406.10721}, 2024.

\bibitem[Yue et~al.(2024)Yue, Ni, Zhang, Zheng, Liu, Zhang, Stevens, Jiang, Ren, Sun, et~al.]{yue2024mmmu}
Xiang Yue, Yuansheng Ni, Kai Zhang, Tianyu Zheng, Ruoqi Liu, Ge~Zhang, Samuel Stevens, Dongfu Jiang, Weiming Ren, Yuxuan Sun, et~al.
\newblock Mmmu: A massive multi-discipline multimodal understanding and reasoning benchmark for expert agi.
\newblock In \emph{Proceedings of the IEEE/CVF Conference on Computer Vision and Pattern Recognition}, pages 9556--9567, 2024.

\bibitem[Zawalski et~al.(2024)Zawalski, Chen, Pertsch, Mees, Finn, and Levine]{zawalski2024robotic}
Micha{\l} Zawalski, William Chen, Karl Pertsch, Oier Mees, Chelsea Finn, and Sergey Levine.
\newblock Robotic control via embodied chain-of-thought reasoning.
\newblock \emph{arXiv preprint arXiv:2407.08693}, 2024.

\bibitem[Zhang et~al.(2024)Zhang, Guo, Chen, Wang, Hu, Shi, and Chen]{zhang2024hirt}
Jianke Zhang, Yanjiang Guo, Xiaoyu Chen, Yen-Jen Wang, Yucheng Hu, Chengming Shi, and Jianyu Chen.
\newblock Hirt: Enhancing robotic control with hierarchical robot transformers.
\newblock \emph{arXiv preprint arXiv:2410.05273}, 2024.

\bibitem[Zhang et~al.(2025)Zhang, Wang, Liu, Huixin, Jiang, Shen, Hou, Zheng, Zhang, Li, et~al.]{zhang2025embodied}
Wenqi Zhang, Mengna Wang, Gangao Liu, Xu~Huixin, Yiwei Jiang, Yongliang Shen, Guiyang Hou, Zhe Zheng, Hang Zhang, Xin Li, et~al.
\newblock Embodied-reasoner: Synergizing visual search, reasoning, and action for embodied interactive tasks.
\newblock \emph{arXiv preprint arXiv:2503.21696}, 2025.

\bibitem[Zheng et~al.(2025{\natexlab{a}})Zheng, Bao, Hou, Zheng, Fang, Huang, Li, Duanmu, Chen, Xu, et~al.]{zheng2025tritondistributed}
Size Zheng, Wenlei Bao, Qi~Hou, Xuegui Zheng, Jin Fang, Chenhui Huang, Tianqi Li, Haojie Duanmu, Renze Chen, Ruifan Xu, et~al.
\newblock Triton-distributed: Programming overlapping kernels on distributed ai systems with the triton compiler, 2025{\natexlab{a}}.
\newblock URL \url{https://arxiv.org/abs/2504.19442}.

\bibitem[Zheng et~al.(2025{\natexlab{b}})Zheng, Fang, Zheng, Hou, Bao, Zheng, et~al.]{zheng2025tilelink}
Size Zheng, Jin Fang, Xuegui Zheng, Qi~Hou, Wenlei Bao, Ningxin Zheng, et~al.
\newblock Tilelink: Generating efficient compute-communication overlapping kernels using tile-centric primitives.
\newblock \emph{arXiv preprint arXiv:2503.20313}, 2025{\natexlab{b}}.

\bibitem[Zhi et~al.(2024)Zhi, Zhang, Zhao, Han, Zhang, et~al.]{zhi2024closed}
Peiyuan Zhi, Zhiyuan Zhang, Yu~Zhao, Muzhi Han, Zeyu Zhang, et~al.
\newblock Closed-loop open-vocabulary mobile manipulation with gpt-4v.
\newblock \emph{arXiv preprint arXiv:2404.10220}, 2024.

\end{thebibliography}
